\documentclass[10pt,journal,compsoc]{IEEEtran}

\ifCLASSOPTIONcompsoc
  \usepackage[nocompress]{cite}
\else
  \usepackage{cite}
\fi

\usepackage{amsmath}
\usepackage{url}
\usepackage[pagebackref=true,breaklinks=true,colorlinks,bookmarks=false, citecolor=black, urlcolor=black, linkcolor=black]{hyperref}
\usepackage{graphicx}
\usepackage{amssymb}
\usepackage{hyperref}
\usepackage{multirow}
\usepackage{caption}
\usepackage{algorithm}
\usepackage{algorithmic}
\usepackage{longtable}
\usepackage{diagbox}
\usepackage{rotating}
\usepackage{color}
\usepackage{makecell}
\graphicspath{{figures/}}

\begin{document}

\title{Rethinking Efficient and Effective  \\Point-based Networks for Event Camera Classification and Regression}

\author{Hongwei Ren, Yue Zhou*, Jiadong Zhu*, Xiaopeng Lin*,Haotian Fu, Yulong Huang, Yuetong Fang,\\Fei Ma ~\IEEEmembership{Member,~IEEE}, Hao Yu, ~\IEEEmembership{Senior Member,~IEEE}, and Bojun Cheng, ~\IEEEmembership{Member,~IEEE}
\IEEEcompsocitemizethanks{
\IEEEcompsocthanksitem Hongwei Ren, Yue Zhou, Jiadong Zhu, Haotian Fu, Yulong Huang, Xiaopeng Lin, Yuetong Fang, and Bojun Cheng are with the MICS Thrust at the Hong Kong University of Science and Technology (Guangzhou). (Corresponding author: Bojun Cheng.) * equal contribution.
\IEEEcompsocthanksitem Fei Ma is in Guangdong Laboratory of Artificial Intelligence and Digital Economy (SZ).

\IEEEcompsocthanksitem Hao Yu is with the School of Microelectronics at Southern University of Science and Technology.
}
}

\markboth{Journal of \LaTeX\ Class Files,~Vol.~14, No.~8, June~2022}%
{Hui \MakeLowercase{\textit{et al.}}: LRRNet: A Novel Representation-learning Guided fusion network for infrared and visible images}
%

\IEEEtitleabstractindextext{
\begin{abstract}
Event cameras draw inspiration from biological systems, boasting low latency and high dynamic range while consuming minimal power. 
The most current approach to processing Event Cloud often involves converting it into frame-based representations, which neglects the sparsity of events, loses fine-grained temporal information, and increases the computational burden. 
In contrast, Point Cloud is a popular representation for processing 3-dimensional data and serves as an alternative method to exploit local and global spatial features.
Nevertheless, previous point-based methods show an unsatisfactory performance compared to the frame-based method in dealing with spatio-temporal event streams.
In order to bridge the gap, we propose EventMamba, an efficient and effective framework based on Point Cloud representation by rethinking the distinction between Event Cloud and Point Cloud, emphasizing vital temporal information.
The Event Cloud is subsequently fed into a hierarchical structure with staged modules to process both implicit and explicit temporal features.
Specifically, we redesign the global extractor to enhance explicit temporal extraction among a long sequence of events with temporal aggregation and State Space Model (SSM) based Mamba.
Our model consumes minimal computational resources in the experiments and still exhibits SOTA point-based performance on six different scales of action recognition datasets. It even outperformed all frame-based methods on both Camera Pose Relocalization (CPR) and eye-tracking regression tasks. Our code is available at: \url{https://github.com/rhwxmx/EventMamba}.
\end{abstract}

}

\maketitle

\IEEEdisplaynontitleabstractindextext

%
\IEEEpeerreviewmaketitle

\IEEEraisesectionheading{\section{Introduction}\label{sec:introduction}}
\IEEEPARstart{E}{vent} cameras present an attractive option for deploying lightweight networks capable of real-time applications.
Unlike traditional cameras, which sample light intensity at fixed intervals, event pixels independently respond to in-situ gradient light intensity changes, indicating the corresponding point's movement in three-dimensional space \cite{lichtsteiner2008128}. This unique capability allows event cameras to capture sparse and asynchronous events that respond rapidly to changes in brightness. Event cameras demonstrate remarkable sensitivity to rapid light intensity changes, detecting log intensity in individual pixels and producing event outputs typically on the order of a million events per second (eps).
In comparison, standard cameras can only capture 120 frames per second, underscoring the exceptional speed of event cameras \cite{yao2021temporal}. Furthermore, the distinctive operating principle of event cameras grants them superior power efficiency, low latency, less redundant information, and high dynamic range. Consequently, event cameras have found diverse applications in high-speed target tracking, simultaneous localization and mapping (SLAM), as well as in industrial automation.

\begin{figure}[t]
\centerline{\includegraphics[width= 7cm]{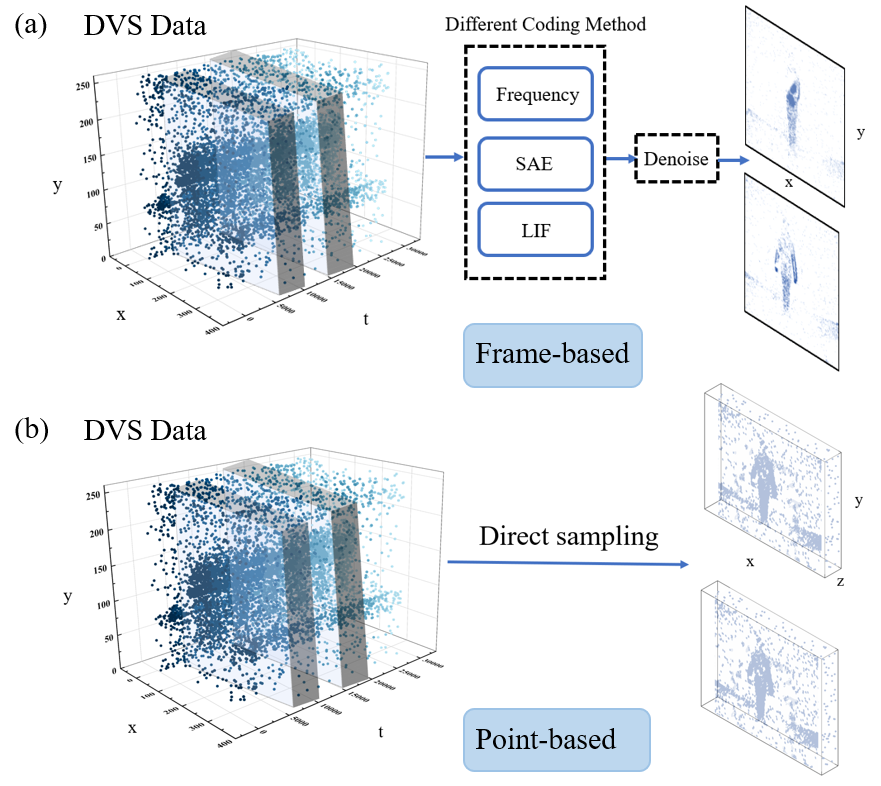}}
\caption{
There are two distinct approaches for managing Event Clouds: $(a)$ frame-based methods and $(b)$ point-based methods. Frame-based techniques involve the condensation of a temporal span of events into a grayscale image using techniques such as SAE, LIF, and others \cite{miao2019neuromorphic}.
}
\label{fig: two method comparison}
\end{figure}

The event camera’s output contains coordinates, timestamps, and polarity of each event occurrence \cite{posch2014retinomorphic}, which is defined as Event Cloud $(x,y,t,p)$. 
The temporal stream can discern fast behaviors with microsecond resolution \cite{gallego2020event}.  Several studies have strived to augment and refine the problem of processing dynamic motion information from event cameras. Figure \ref{fig: two method comparison} (a) illustrates the frame-based method, which partitions the Event Cloud into small chunks and accumulates the events for each time interval onto the image plane \cite{gehrig2019end,lin2020efficient}. The grayscale on the plane denotes the intensity of the event occurrence at that coordinate. However, the frame-based method disregards the fine-grained temporal information within a frame and may generate blurred images in fast-moving action scenarios. Moreover, this approach escalates the data density and computes even in event-free regions, forfeiting the sparse data benefits of an event-based camera. Consequently, the frame-based method is unsuitable for scenarios with fast-moving actions and cannot be implemented with limited computational resources.

\begin{figure*}[t]
\centerline{\includegraphics[width= 18 cm]{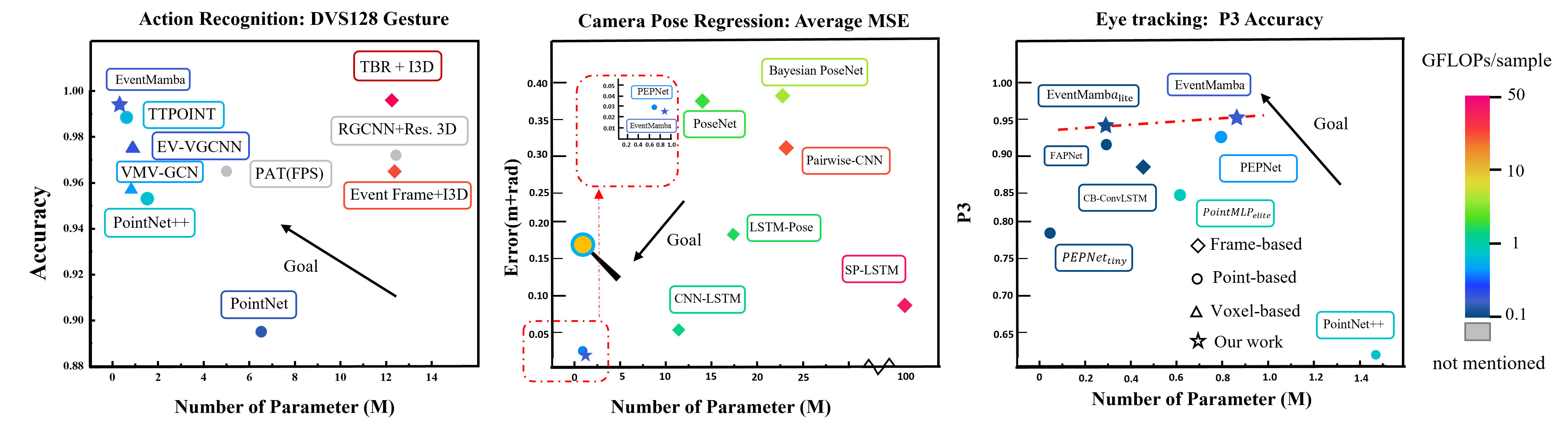}}
\caption{The gain in three datasets is visualized. EventMamba achieves SOTA in the point-based method and maintains very high efficiency. The shapes illustrate the results corresponding to different data formats, with varying colors denoting Floating Point Operations (FLOPs).}
\label{fig: gain}
\end{figure*}

In order to overcome the issues discussed above, alternative approaches, such as the point-based method have been proposed \cite{wang2019space,ren2023ttpoint,ren2024simple}. Point Cloud is a collection of 3D points $(x,y,z)$ that represents the shape and surface of an object or environment and is often used in lidar and depth cameras \cite{guo2020deep}.  
By treating each event's temporal information as the third dimension, event inputs $(x,y,t)$ can be regarded as points and aggregated into a pseudo-Point Cloud \cite{wang2019space,qi2017pointnet,qi2017pointnet++}. This method is well-suited for analyzing Event Cloud, which can then be directly fed into the Point Cloud network, as shown in Figure \ref{fig: two method comparison} (b). By doing so, the data format transformation process is greatly simplified, and the fine-grained temporal information and sparse property are preserved \cite{wang2019space}.

Nonetheless, contemporary point-based approaches still exhibit a discernible disparity in performance compared to frame-based approaches. 
Although the Event Cloud and the Point Cloud are identical in shape, the information they represent is inconsistent, as mentioned above. Merely transplanting the most advanced Point Cloud network proves insufficient in extracting Event Cloud features, as the permutation invariance and transformation invariance inherent in Point Cloud do not directly translate to the domain of Event Cloud \cite{qi2017pointnet++}. 
Meanwhile, Event Cloud encapsulates a wealth of temporal information vital for event-based tasks. Consequently, addressing these distinctions necessitates the development of a tailored neural network module to effectively leverage the temporal features $t$ within the Event Cloud. This requests not only implicit temporal features, wherein $t$ and $x,y$ are treated as the same representation, but also explicit temporal features, where $t$ serves as the index of the features.

In this paper, we introduce a novel Point Cloud framework, named EventMamba, for event-based action recognition, CPR, and eye-tracking tasks, whose gains are shown in Figure \ref{fig: gain}. Consistency with our previous works\cite{ren2023ttpoint,ren2024simple}, EventMamba is characterized by its lightweight parameterization and computational demands, making it ideal for resource-constrained devices. Diverging from our previous works, TTPOINT \cite{ren2023ttpoint}, which did not consider temporal feature extraction, and PEPNet \cite{ren2024simple}, which employed A-Bi-LSTM behind the hierarchical structure, we try to integrate the explicit temporal feature extraction module into the hierarchical structure to enhance temporal features, forming a unified, efficient, and effective framework. We also embrace the hardware-friendly Mamba Block, which demonstrates proficiency in processing sequential data.  
The main contributions of our work are: 
\begin{itemize}
    \item We directly process the raw data obtained from the event cameras, preserving the sparsity and fine-grained temporal information.
    \item We strictly arrange the events in chronological order, which can satisfy the prerequisites to extract the temporal features absent in the vanilla Point Cloud Network.
    \item  We employ temporal aggregation and Mamba block to enhance explicit temporal features during the hierarchy structure effectively.
    \item We propose a lightweight framework with a generalization ability to handle different-size event-based classification and regression tasks.
\end{itemize}

\section{Related Works}
\label{sec:related}
\subsection{Frame-based Method}
Frame-based methods have been extensively employed for event data processing. 
The following will offer a comprehensive review of frame-based methods applied to three distinct tasks.
\subsubsection{Action Recognition}
Action recognition is a critical task with diverse applications in anomaly detection, entertainment, and security. Most event-based approaches often have borrowed from traditional camera-based approaches. The integration of the number of events within time intervals at each pixel to form a visual representation of a scene has been extensively studied, as documented in \cite{innocenti2021temporal}.  In cases where events repeatedly occur at the same spatial location, the resulting frame information is accumulated at the corresponding address, thus generating a grayscale representation of the temporal evolution of the scene. This concept is further explored in related literature, exemplified by the Event Spike Tensor (EST) \cite{gehrig2019end} and the Event-LSTM framework \cite{annamalai2022event}, both of which demonstrate methods for converting sparse event data into a structured two-dimensional temporal-spatial representation, effectively creating a time-surface approach.

The resultant 2D grayscale images, encoding event intensity information, can be readily employed in traditional feature extraction techniques. Notably, Amir et al. \cite{amir2017low} directly fed stacked frames into Convolutional Neural Networks (CNN), affirming the feasibility of this approach in hardware implementation. Bi et al. \cite{bi2020graph} introduced a framework that utilizes a Residual Graph Convolutional Neural Network (RG-CNN) to learn appearance-based features directly from graph representations, thus enabling an end-to-end learning framework for tasks related to appearance and action. Deng et al. \cite{deng2021learning} proposed a distillation learning framework, augmenting the feature extraction process for event data through the imposition of multiple layers of feature extraction constraints.

\subsubsection{Camera Pose Relocalization}
CPR task is an emerging application in power-constrained devices and has gained significant attention. It aims to train several scene-specific neural networks to accurately relocalize the camera pose within the original scene used for training \cite{kendall2015posenet}. 
Event-based CPR methods often derive from the frame-based CPR network \cite{shavit2019introduction,walch2017image,naseer2017deep,walch2017image,wu2017delving,naseer2017deep,brachmann2021visual}. SP-LSTM \cite{nguyen2019real} employed the stacked spatial LSTM networks to process event images, facilitating a real-time pose estimator. To address the inherent noise in event images, Jin Yu et al.\cite{jin20216} proposed a network structure combining denoise networks, convolutional neural networks, and LSTM, achieving good performance under complex working conditions. In contrast to the aforementioned methods, a novel representation named Reversed Window Entropy Image (RWEI) \cite{lin20226} is introduced, which is based on the widely used event surface \cite{mitrokhin2020learning} and serves as the input to an attention-based DSAC* pipeline \cite{brachmann2021visual} to achieve SOTA results. However, the computationally demanding architecture involving representation transformation and hybrid pipeline poses challenges for real-time execution. Additionally, all existing methods ignore the fine-grained temporal feature of the event cameras, and accumulate events into frames for processing, resulting in unsatisfactory performance.

\subsubsection{Eye-tracking}
Event-based eye-tracking methods leverage the inherent attribution of event data to realize low-latency tracking. Ryan et al. \cite{ryan2021real} propose a fully convolutional recurrent YOLO architecture to detect and track faces and eyes. Stoffregen et al. \cite{stoffregen2022event} design Coded Differential Lighting method in corneal glint detection for eye tracking and gaze estimation.  Angelopoulos et al. \cite{angelopoulos2021event} propose a hybrid frame-event-based near-eye gaze tracking system, which can offer update rates beyond 10kHz tracking. Zhao et al. \cite{zhao2024ev} also leverage the near-eye grayscale images with event data for a hybrid eye-tracking method. The U-Net architecture is utilized for eye segmentation, and a post-process is designed for pupil tracking. Chen et al. \cite{chen20233et} presents a sparse Change-Based Convolutional Long Short-Term Memory model to achieve high tracking speed in the synthetic event dataset. 

In summary, frame-based methods have indeed showcased their prowess in event processing tasks, but they suffer from several limitations. For instance, the processed frame sizes are typically larger than those of the original Event Cloud, resulting in high computational costs and large model sizes. Furthermore, the aggression process introduces significant latency \cite{gehrig2019end}, which hinders the use of event cameras in real-time human-machine interaction applications. Despite substantial endeavors aimed at mitigating these challenges \cite{zubic2024state}, changing the representation stands as the suitable approach to delving deep into the core of the issue.

\begin{figure*}[t]
\centerline{\includegraphics[width= 18 cm]{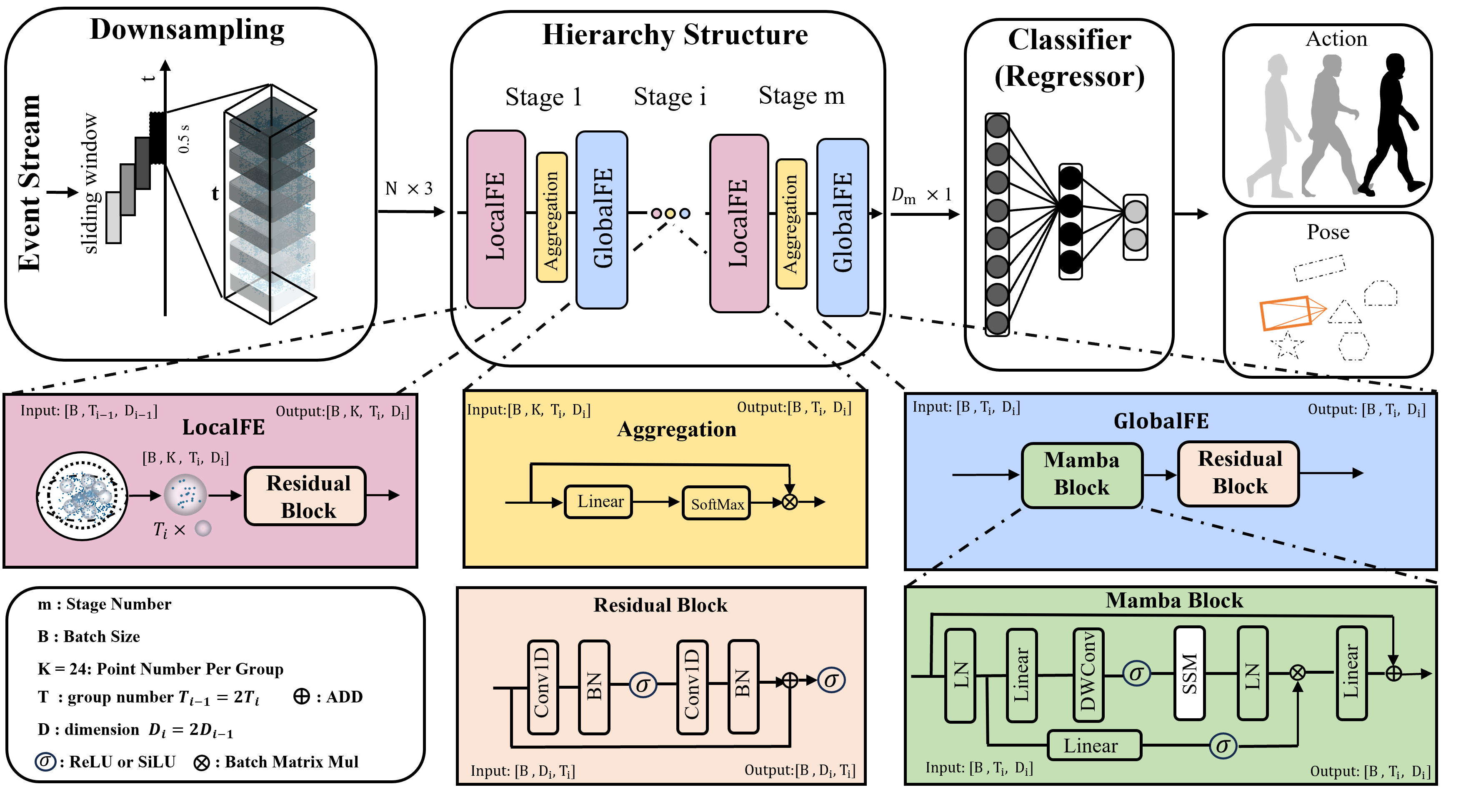}}
\caption{
EventMamba accomplishes two tasks by processing the Event Cloud to a sequence of distinct modules, downsampling, hierarchy structure, and classifier or regressor. More specifically, the $LocalFE$  is responsible for the extraction of local geometric features, while the $GlobalFE$  plays a pivotal role in elevating the dimensionality of the extracted features and abstracting higher-level global and explicit temporal features.
}
\label{fig: eventmamba arch}
\end{figure*}
\subsection{Point-based Method}
With the introduction of PointNet \cite{qi2017pointnet}, Point  Cloud can be handled directly as input, marking a significant evolution in processing technologies. This development is further enhanced by PointNet++ \cite{qi2017pointnet++}, which integrates the Set Abstraction (SA) module. This module serves as an innovative tool to elucidate the hierarchical structure related to global and local Point Cloud processing. Although the original PointNet++ relies on basic MLPs for feature extraction, there have been considerable advancements to enhance Point Cloud processing methodologies \cite{wu2019pointconv,zhao2021point,liang2024pointmamba,zhang2024point,liu2024point}. 
Furthermore, an intriguing observation is that even simplistic methodologies can yield remarkable outcomes. A case in point is PointMLP \cite{ma2022rethinking}, despite its reliance on an elementary ResNet block and affine module, consistently achieves SOTA results in the domains of point classification and segmentation.

Utilizing point-based methods within the Event Cloud necessitates the handling of temporal data to preserve its integrity alongside the spatial dimensions \cite{sekikawa2019eventnet}. The pioneering effort in this domain was manifested by ST-EVNet \cite{wang2019space}, which adapted PointNet++ for the task of gesture recognition. Subsequently, PAT \cite{yang2019modeling} extended these frontiers by integrating a self-attention mechanism with Gumbel subset sampling, thereby enhancing the model's performance within the DVS128 Gesture dataset. Recently, TTPOINT \cite{ren2023ttpoint} utilized the tensor train decomposition method to streamline the Point Cloud Network and achieve satisfied performance on different action recognition datasets. However, these methods rely on the direct transplant of the Point Cloud network, which is not adapted to the explicit temporal characteristics of the Event Cloud. Moreover, most models are not optimized for size and computational efficiency and cannot meet the demands of edge devices \cite{xie2022vmv}.
\subsection{State Space Models}
State Space Models (SSM) have emerged as a versatile framework initially rooted in control theory, now widely applied due to its superior capability process sequence modeling tasks \cite{gu2021efficiently,gupta2022diagonal,gu2022parameterization}. 
SSM-based Mamba \cite{gu2023mamba} has shown promise as a hardware-friendly backbone that can achieve linear-time inference and can be an alternative to the transformer. While primarily focused on language and speech data, recent advancements like S4ND \cite{nguyen2022s4nd}, Vision Mamba \cite{zhu2024vision}, and PointMamba \cite{liang2024pointmamba} have extended SSMs to visual tasks, showing potential in competing with established models like ViTs. Nevertheless, neither the image nor Point Cloud contains strict temporal information, requiring both to employ customized serialization and ordering methods to align with the SSM sequences \cite{wang2024state}. Event-based works \cite{zubic2024state,huang2024mamba} also focus on frame-based representations, which do not take full advantage of the rich and fine-grained temporal information in events. In this paper, we leverage point cloud representations, preserving the temporal sequence of event occurrences, and integrated Mamba with a hierarchical structure to comprehensively extract temporal features between events.

\section{Methodology}
\subsection{Event Cloud Preprocessing}
To preserve the fine-grained temporal information and original data distribution attributes from the Event Cloud, the 2D-spatial and 1D-temporal event information is constructed into a three-dimensional representation to be processed in the Point Cloud. Event Cloud consists of time-series data capturing spatial intensity changes of images in chronological order, and an individual event is denoted as $e_k=(x_k, y_k, t_k, p_k)$, where $k$ is the index representing the $k_{th}$ element in the sequence. Consequently, the set of events within a single sequence ($\mathcal{E}$) in the dataset can be expressed as:
\begin{equation}
        \mathcal{E} = \left\{e_k=(x_k,y_k,t_k,p_k) \mid k=1, \ldots, n\right\},
\end{equation}
The length of the Event Cloud in the dataset exhibits variability. To ensure the consistency of inputs to the network, we employ a sliding window to segment the Event Cloud. In the task of action recognition, it is imperative to include a complete action within a sliding window. 
Therefore, we choose to set the sliding window length in seconds.
Meanwhile, for the task of Camera Pose Relocalization and eye tracking, the resolution of the pose ground truth is available at the millisecond level. 
As a result, we chose to set the sliding window length to milliseconds.
The splitting process can be represented by the following equation:
\begin{equation}
    P_i = \{e_{j \rightarrow l} \mid t_l - t_j = R \}, \quad i=1,\ldots,L
\end{equation}
The symbol $R$ represents the time interval of the sliding window, where $j$ and $l$ denote the start and end event index of the sequence, respectively. $L$ represents the number of sliding windows into which the sequence of events $\mathcal{E}$ is divided. 
Before being fed into the neural network, $P_i$ also needs to undergo downsampling and normalization. Downsampling is to unify the number of Point Cloud, with the number of points in a sample defined as $N$, which is 1024 or 2048 in EventMamba. Additionally, we define the resolution of the event camera as $w$ and $h$. The normalization process is described by the following equation:
\begin{equation}
    PN_i = ( \frac{X_i}{w},\frac{Y_i}{h},\frac{TS_{i}-t_j}{t_l - t_j}),  
\end{equation}
\begin{equation}
    X_i, Y_i, TS_{i} = \{x_j, \dots, x_l\}, \{y_j, \dots, y_l\}, \{t_j, \dots, t_l\},
\end{equation}
The $X, Y$ is divided by the resolution of the event camera. To normalize timestamps $TS$, we subtract the smallest timestamp $t_j$ of the window and divide it by the time difference $t_l - t_j$, where $t_l$ represents the largest timestamp within the window. After pre-processing, Event Cloud is converted into the pseudo-Point Cloud, which comprises explicit spatial information $(x,y)$ and implicit temporal information $t$. It is worth mentioning that all events in $PN$ are strictly ordered according to the value of $t$, ensuring that the network can explicitly capture time information.

\subsection{Hierarchy Structure}
In this section, we provide a detailed exposition of the network architecture and present a formal mathematical representation of the model. 
Once we have acquired a sequence of 3D pseudo-Point Clouds denoted as $PN$, with dimensions of $[L, N,3]$, through the aforementioned pre-processing method applied to the Event Cloud, we are prepared to input this data into the hierarchy structure model for the purpose of extracting Point Cloud features.
We adopt the residual block in the event processing field, which is proven to achieve excellent results in the Point Cloud field \cite{ma2022rethinking}. Next, we will introduce the pipeline for the hierarchy structure model.
\begin{align}
&\mathcal{S}_i = LocalFE(\mathcal{ST}_{i-1}),\\
&\mathcal{SA}_i = TAG(\mathcal{S}_i),\\
&\mathcal{ST}_i = GlobalFE(\mathcal{SA}_i), \\
F = \mathcal{ST}_m,\quad &\mathcal{ST}_0=PN_i, \quad where\quad i \in (1,m), 
\end{align}
where $F$ is the extracted spatio-temporal Event Cloud feature, which can be sent into the classifier or regressor to complete the different tasks. The model has the ability to extract global and local features and has a hierarchy structure \cite{qi2017pointnet++}. And the dimension of $LocalFE$ input $\in$ is $[B, T_{i-1}, D_{i-1}]$, $m$ is the number of stages for hierarchy structure, $TAG$ is the operation of temporal aggregation, $\mathcal{S}$ is the output of $LocalFE$, and the dimension of $GlobalFE$ input $\in$ $[B, T_i, D_i]$, as illustrated in Figure \ref{fig: eventmamba arch}. $B$ is the batch size, $D$ is the dimension of features, and $T$ is the number of $Centroid$ in the overall Point Cloud during the current stage of the hierarchy structure. 

From a macro perspective, $LocalFE$ is mainly used to extract the local features of the spatial domain and pay attention to local geometric features
through the farthest point sampling and grouping method. Temporal aggregation ($TAG$) is responsible for aggregating local features on the timeline. $GlobalFE$ mainly synthesizes the extracted local features and captures the explicit temporal relationship between different groups. Details of the three modules are described below.
\subsubsection{$LocalFE$ Extractor}
This extractor comprises a cascade of sampling, grouping, and residual blocks. Aligned with the frame-based design concept \cite{yuan2024learning}, our focus is to capture both local and global information. Local information is acquired by leveraging Farthest Point Sampling (FPS) to sample and K-nearest neighbors (KNN) to the group. The following equations can represent the operations of sampling and grouping.
\begin{equation}
    PS_{i}=FPS(PN_i), \quad
    PG_{i}=KNN(PN_i,PS_{i},K),
\end{equation}
The input dimension $PN_i$ is $[T,3+D]$, and the centroid dimension $PS_{i}$ is $[T^{'},3+D]$ and the group dimension $PG_{i}$ is $[T^{'}, K,3+2*D]$. $K$ represents the nearest $K$ points of centroid, $D$ is the feature dimension of the points of the current stage, and 3 is the most original $(X, Y, TS)$ coordinate value. $2*D$ is to concatenate the features of the point with the features of the centroid.

Next, each group undergoes a standardization process to ensure consistent variability between points within the group, as illustrated in this formula:
\begin{equation}
    PGS_i=\frac{PG_i-PS_i}{Std(PG_i)}, \quad \text{Std}(PG_i)=\sqrt{\frac{ {\textstyle \sum_{i=0}^{3n-1}} (g_i - \bar g)^2}{3n-1} },
\end{equation}
\begin{equation}
    g =[x_0,y_0,t_0,\dots,x_n,y_n,t_n],
\end{equation}
Where $\text{Std}$ is the standard deviation, and $g$ is the set of coordinates of all points in the $PG_i$.

Subsequently, the residual block is utilized to extract higher-level features.
Concatenating residual blocks can indeed enhance the model's capacity for learning, yet when dealing with sparsely sampled event-based data, striking the right balance between parameters, computational demands, and accuracy becomes a pivotal challenge. In contrast to frame-based approaches, the strength of Point Cloud processing lies in its efficiency, as it necessitates a smaller dataset and capitalizes on the unique event characteristics. However, if we were to design a point-based network with parameters and computational demands similar to frame-based methods, we would inadvertently forfeit the inherent advantages of leveraging Point Clouds for event processing. The Residual block (\text{ResB}) can be written as a formula:
\begin{align}
&M(x) = BN(MLP(x)), \\
&\text{ResB}(x)= \sigma(x+M(\sigma(M(x)))), 
\end{align}
 where $\sigma$ is a nonlinear activation function ReLU, $x$ is the input features, $BN$ is batch normalization, and $MLP$ is the muti-layer perception module.
\subsubsection{Temporal Aggregation}
Following the processing of points in a group by $LocalFE$ extractor, the resulting features require integration via the aggregation module. This module is tailored to necessitate precise temporal details within a group. Consequently, we introduce a temporal aggregation module.

\begin{figure}[t]
\centerline{\includegraphics[width= 7cm]{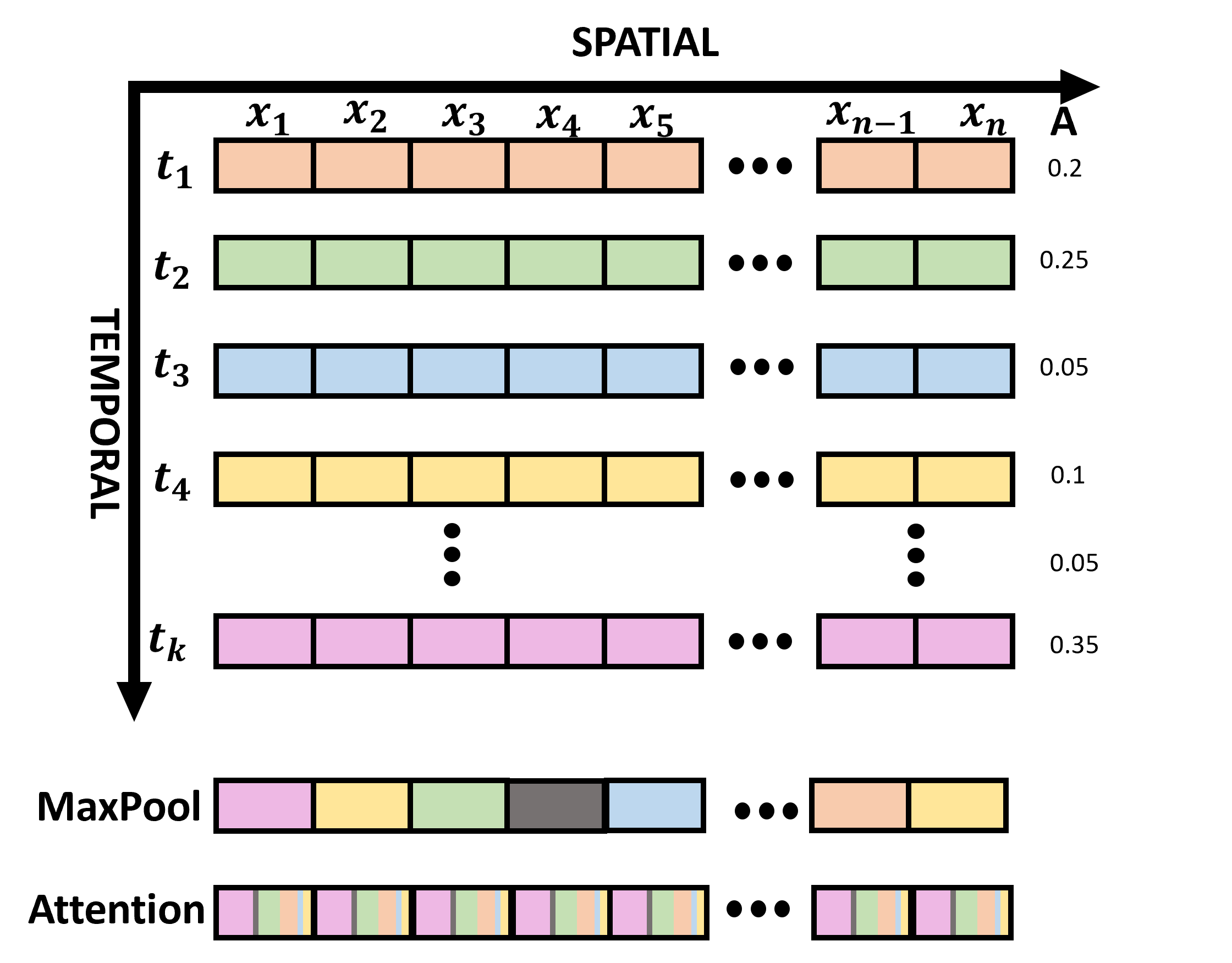}}
\caption{Spatial and temporal aggregation. The different colors represent the features of events at different times, with gray representing the feature in $[t_5,t_{k-1}]$.}
\label{fig: aggregation}
\end{figure}
Conventional Point Cloud methods favor MaxPooling operations for feature aggregation because it is efficient in extracting the feature from one point among a group of points and discarding the rest. In other words, MaxPooling involves extracting only the maximum value along each dimension of the temporal axis. It is robust to noise perturbation but also ignores the temporal nuances embedded within the features. This property matches well with the vanilla Point Cloud task, which only needs to focus on the most important information. However, the event-based task requires fine-grained temporal features, so we adopt the integration of attention mechanisms to enhance the preservation of those nuanced but useful temporal attributes by aggregating features along the temporal axis through the attention value. To provide a more comprehensive exposition, we employ a direct attention mechanism within the $K$ temporal 
dimensions to effectively aggregate features as shown in Figure \ref{fig: aggregation}. This mechanism enables the explicit integration of temporal attributes, capitalizing on the inherent strict ordering of the $K$ points. The ensuing formula succinctly elucidates the essence of this attention mechanism:
\begin{equation}
    \mathcal{S} = LocalFE(x) = (S_{t1},S_{t2},\dots,S_{tk}),
\end{equation}
\begin{equation}
    S_{tk} = [x_1,x_2,\dots,x_n],
\end{equation}
\begin{equation}
    A = \text{SoftMax}(MLP(\mathcal{S})) = (a_{t1},a_{t2},\dots,a_{tk}),
\end{equation}
\begin{equation}
    \mathcal{SA} = \mathcal{S} \cdot A = S_{t1}\cdot a_{t1} + S_{t2}\cdot a_{t2} + \dots + S_{tk}\cdot a_{tk},
\end{equation}
Upon the application of the local feature extractor, the ensuing features are denoted as $\mathcal{S}$, and $S_{tk}$ mean the extracted feature of $k_{th}$ point in a group. The attention mechanism comprises an MLP layer with an input layer dimension of $D$ and an output $a_{tk}$ dimension of 1, along with softmax layers. Subsequently, the attention mechanism computes attention values, represented as $A$.  These attention values are then multiplied with the original features through batch matrix multiplication, resulting in the aggregated feature $\mathcal{SA}$.

\subsubsection{$GlobalFE$ Extractor} Unlike previous work\cite{ren2024simple},  employing Bi-LSTM for explicit temporal information extraction nearly doubles the number of parameters, despite achieving amazing results. We incorporate the explicit temporal feature extraction module into the hierarchical structure to enhance temporal features, opting for the faster and stronger Mamba block instead \cite{gu2023mamba}. The topology of the block is shown in the green background of Figure \ref{fig: eventmamba arch}. This block mainly integrates the sixth version of the State Space Model (SSM), which is well able to parallel and focus on long time series of information, such as temporal correlations between 1024 event groups. The SSM extracts explicit temporal features that can be expressed as follows:
\begin{align}
    h_t = & \bar{\mathbf{A}}h_t + \bar{\mathbf{B}}x_t,  \quad t \in [0,T') \quad h\in D'\\
    y_t = & \mathbf{C}h_t, \quad y\in D'\\
    \bar{\mathbf{A}} = & exp(\Delta \mathbf{A}), \quad \bar{\mathbf{A}},\bar{\mathbf{B}}:(T',D',D')\\
    \bar{\mathbf{B}}=&(\Delta \mathbf{A})^{-1}(\exp (\Delta \mathbf{A})-I) \cdot \Delta \mathbf{B},  
\end{align}
Where $x_t,h_t,$ and $y_t$ are the SSM's discrete inputs, states, and outputs. $\mathbf{A}$, $\mathbf{B}$ and $\mathbf{C}$ are the continuous system parameters, while $\bar{\mathbf{A}}$, $\bar{\mathbf{B}}$ and $\Delta$ are the parameters in the discrete system by the zero-order hold rule. $T'$ and $D'$ are the number and dimension of events for the current stage, respectively.
The whole $GlobalFE$ Extractor can be represented by this formula:
\begin{align}
    \mathcal{ST} = & \quad \text{Mamba}(\mathcal{SA}) \\
    \mathcal{ST} = & \quad \text{ResB}(\mathcal{ST})
\end{align}
Where Mamba extracts explicit temporal features in the $T'$ dimension, while ResB further abstracts the spatial and temporal $\mathcal{ST}$ features.

\subsection{Regressor and Loss Function}
 A fully connected layer with a hidden layer is employed to address the final classification and regression task. \textbf{Classification task:} We employ label smoothing to refine the training process. Label smoothing involves introducing slight smoothing into the true label distribution, aiming to prevent the model from becoming overly confident in predicting a single category. The label-smoothed cross-entropy loss function we define is as follows:
 \begin{equation}
     Loss = -\frac{1}{n} \sum_i ( (1 - \epsilon) \log(\hat{y}_i) + \alpha \sum_{j \neq i} \log\left(\alpha\right) )
 \end{equation}
$n$ represents the number of categories, $\epsilon$ is the smoothing parameter, $\alpha$ represents $\frac{\epsilon}{n-1}$, and $\hat{y}_i$ denotes the predicted probability of category $i$ by the model. \textbf{Camera Pose Relocalization task:} The displacement vector of the regression is denoted as $\hat{p}$ representing the magnitude and direction of movement, while the rotational Euler angles are denoted as $\hat{q}$ indicating the rotational orientation in three-dimensional space.
\begin{equation}
    Loss = \alpha ||\hat p - p ||_2 +\beta ||\hat q -q||_2+\lambda  {\textstyle \sum_{i=0}^{n}}  w_i^2
\end{equation}
$p$ and $q$ represent the ground truth obtained from the dataset, while $\alpha$, $\beta$, and $\lambda$ serve as weight proportion coefficients. In order to tackle the prominent concern of overfitting, especially in the end-to-end setting, we propose the incorporation of L2 regularization into the loss function. This regularization, implemented as the second paradigm for the network weights $w$, effectively mitigates the impact of overfitting. \textbf{Eye tracking task:} The weighted Mean Squared Error (WMSE) is selected as the loss function to constrain the distance between the predicted pupil location and the ground truth label to guide the network training. The loss function is shown as follows:
\begin{align}
\label{eq:wmse}
Loss  = \nonumber w_x \cdot \frac{1}{n} \sum_{i=1}^{n} (\hat{x} - x)^2  + w_y \cdot \frac{1}{n} \sum_{i=1}^{n} ( \hat{y} - y)^2,
\end{align}
where $\hat{x}$ and $\hat{y}$ stand for the prediction value along the $x$ and $y$ dimensions. $x$ and $y$ are the ground truth values along the x and y dimensions. $w_x$ and $w_y$ are the hyperparameters to adjust the weight of the x dimension and y dimension in the loss function. $i$ means the $i_{th}$ pixel of the prediction or ground truth label.
\subsection{Overall Architecture}
Next, we will present the EventMamba pipeline in pseudo-code, utilizing the previously defined variables and formulas as described in Algorithm 1.
\begin{algorithm}[h]
\caption{EventMamba Pipeline}
\textbf{Input}: Raw Event Cloud $\mathcal{E}$\\
\textbf{Parameters}: Stage Number: $m$, Sliding window threshold: $N$ and time interval: $R$\\
\textbf{Output}: Category $\hat{c}$, Position $(\hat{p},\hat{q})$, $(\hat{x},\hat{y})$

\begin{algorithmic}[1] 
\STATE\textbf{Event Cloud Preprocessing}
\FOR{ $j$ \textbf{in} len($\mathcal{E}$)}
\STATE $P_i.\text{append}(e_{j\rightarrow l})$ ; $j=l$; where $t_l-t_j=R$
\STATE \textbf{if} ($\text{len}(P_i)>N$): $i=i+1$;
\ENDFOR
\STATE $PN = \text{Normalize}(\text{Downsample}(P))$;
\STATE \textbf{Hierarchy Structure}
\FOR{ i in $\text{stage}(m)$}
\STATE \textbf{Local Extractor}: Get $\mathcal{S} \in [B,T_{i-1},K,D_{i-1}]$
\STATE $\mathcal{S} = LocalFE(PN)$;
\STATE \textbf{Temporal Aggregate}: Get $\mathcal{SA} \in [B,T_{i},D_{i}]$
\STATE $\mathcal{SA} = TAG(\mathcal{S})$;
\STATE \textbf{Global Extractor}: Get $PN\in [B, T_{i}, D_{i}]$
\STATE $PN= GlobalFE(\mathcal{SA})$;
\ENDFOR
\STATE \textbf{Classifier } 
\STATE Get category $\hat{c}$
\STATE \textbf{Regressor } 
\STATE Get 6-DOFs pose and eye coordinate
\end{algorithmic}
\end{algorithm}
\section{EXPERIMENTS}
\label{EXPERIMENTS}
In this section, we train and test the performance of EventMamba on the server platform. The experimental evaluation encompasses assessing the model's accuracy, mean square error (MSE), and pixel accuracy performance on the datasets, analyzing the model's parameter size, inference time, and the number of floating-point operations (FLOPs). The platform configuration comprises the following: CPU: AMD 7950x, GPU: RTX 4090, and Memory: 32GB.
\subsection{Datasets}
\begin{figure*}[t]
\centering
\centerline{\includegraphics[width=18cm]{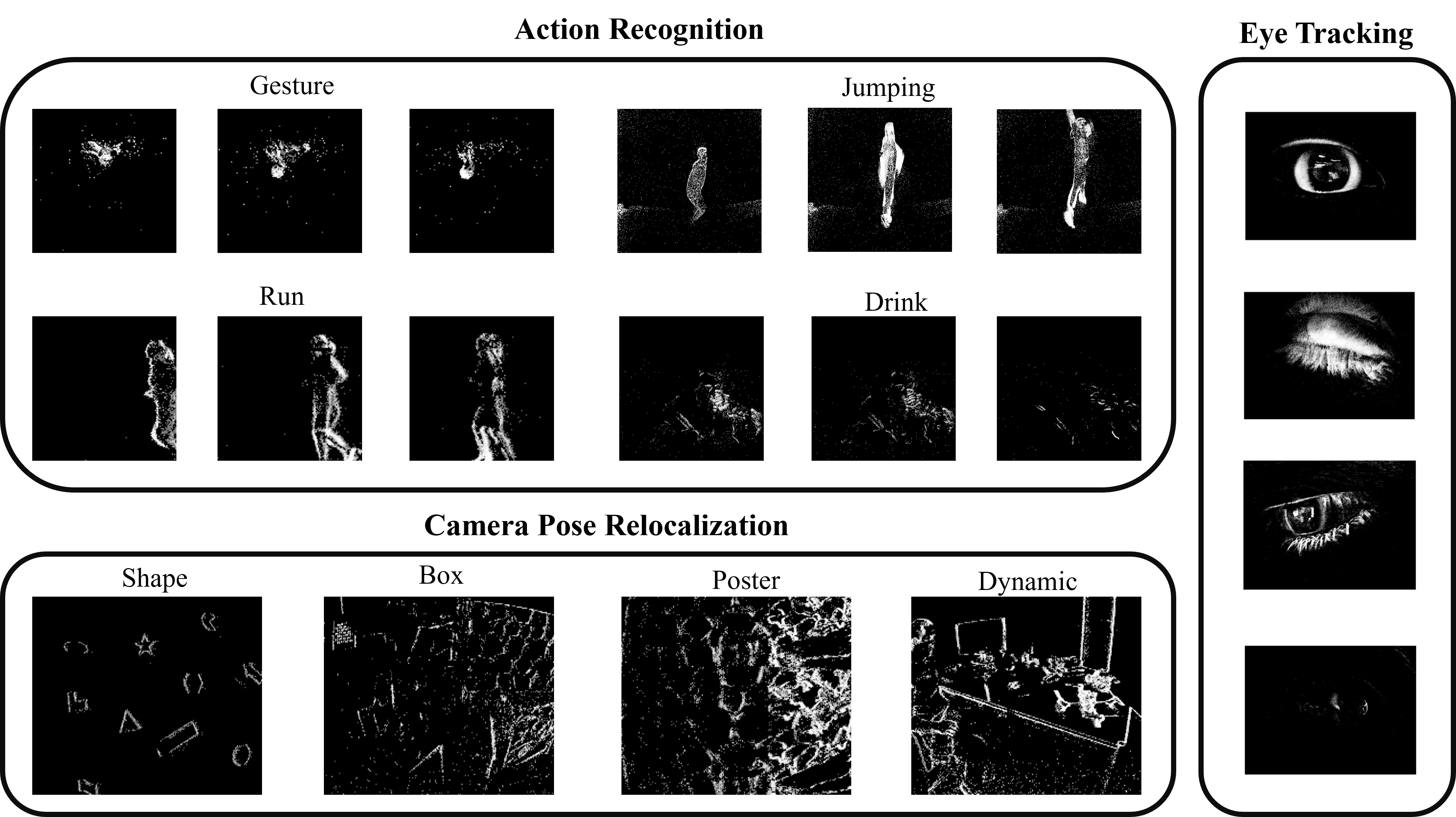}}
\caption{Event-based action recognition, CPR and eye tracking datasets visualization. 
}
\label{fig}
\label{fig: visualize dataset}
\end{figure*}
We evaluate EventMamba on six popular action recognition datasets, the IJRR Camera Pose Relocalization dataset, and the SEET eye tracking dataset for classification and regression. The resolution, classes, and other information of datasets are shown in Table \ref{table: infor for dataset}.

\textbf{Daily DVS} dataset \cite{liu2021event} consists of 1440 recordings involving 15 subjects engaged in 12 distinct everyday activities, including actions like standing up, walking, and carrying boxes. Each subject performed each action under uniform conditions, with all recordings within a duration of 6 seconds.

\textbf{DVS128 Gesture} dataset \cite{amir2017low} encompasses 11 distinct classes of gestures, which were recorded by IBM in a natural environment utilizing the iniLabs DAVIS128 camera. This dataset maintains a sample resolution of 128$\times$128, signifying that the coordinate values for both x and y fall within the range of [1, 128]. It comprises a total of 1342 examples distributed across 122 experiments that featured the participation of 29 subjects subjected to various lighting conditions.

\textbf{DVS Action} dataset \cite{miao2019neuromorphic} comprises 10 distinct action classes, all captured using the DAVIS346 camera. The sample resolution for this dataset is set at 346$\times$260. The recordings were conducted in an unoccupied office environment, involving the participation of 15 subjects who performed a total of 10 different actions. 

\textbf{HMDB51-DVS and UCF101-DVS} datasets, as detailed in \cite{bi2020graph}, are adaptations of datasets originally captured using conventional cameras, with the conversion process utilizing the DAVIS240 camera model. UCF101 encompasses an extensive compilation of 13,320 videos, each showcasing 101 unique human actions. In contrast, HMDB51 is comprised of 6,766 videos, each depicting 51 distinct categories of human actions. Both datasets maintain a sample resolution of 320$\times$240.

\textbf{THU$^{\text{E-ACT}}$-50-CHL} dataset \cite{gao2023action} is under challenging scenarios.
It comprises 2,330 recordings featuring 18 students, captured from varied perspectives utilizing a DAVIS346 event camera in two distinct scenarios: a long corridor and an open hall. Every recording lasts between 2 to 5 seconds and maintains a spatial resolution of 346 × 260.

\textbf{IJRR CPR} \cite{mueggler2017event} dataset collected using the DAVIS 240C, encompassing both outdoor and indoor scenes. This dataset encompasses a diverse set of multimodal information comprising events, images, Inertial Measurement Unit (IMU) measurements, camera calibration, and ground truth information acquired from a motion capture system operating at an impressive frequency of 200 Hz, thereby ensuring sub-millimeter precision. We visualized various types of sequences as shown in Figure \ref{fig: visualize dataset}.

\textbf{SEET} dataset \cite{chen20233et} is a synthetic eye-tracking dataset simulated from an RGB dataset named Labeled Pupils in the Wild \cite{tonsen2016labelled}. Event streams are generated through the v2e simulator with the $240 \times 180$ resolution\cite{hu2021v2e}.

\begin{table*}
\caption{Information of Different Action Recognition and CPR Datasets with Event Camera.}
\centering
\renewcommand\arraystretch{1}
\scalebox{0.9}{
\begin{tabular}{cccccccccc}
\hline
Dataset            & Classes       & Sensor        & Resolution         & Avg.length(s) & Train Samples & Test Samples &Point Number &Sliding window  &Overlap\\ \hline
Daily DVS \cite{liu2021event}          &12            &  DAVIS128              &  128x128              &3   &2924 &731 &2048 &1.5s &0.5s\\
DVS128 Gesture \cite{amir2017low}   & 10/11         & DAVIS128      & 128x128                 & 6.52          & 26796         & 6959 &1024 &0.5s &0.25s\\
DVS Action \cite{miao2019neuromorphic} & 10            & DAVIS346      & 346x260                & 5             & 932          & 233 &1024 &0.5s &0.25s\\
HMDB51-DVS  \cite{bi2020graph}      & 51            & DAVIS240      & 320x240                 & 8             & 48916         & 12230  &2048 &0.5s &0.25s\\ 
UCF101-DVS  \cite{bi2020graph}       &101            &  DAVIS240             & 320x240             & 6.6  &110050 &27531 &2048 &1s &0.5s\\
THU$^{\text{E-ACT}}$-50-CHL  \cite{gao2023action}       &50            &  DAVIS346             & 346x260             & 2-5  &6528 &2098 &2048 &1s &0.5s\\
IJRR \cite{mueggler2017event} & 6 DOFs & DAVIS240 & 320x240 & 60 & - & - &1024 & 5 ms & 0s \\
SEET \cite{chen20233et} & 2 &DAVIS240 &240x180 &20 &31861 &3973 &1024 &4.4 ms &0s\\\hline
\end{tabular}}
\label{table: infor for dataset}
\end{table*}

\subsection{Implement Details}
\subsubsection{Pre-processing}
The temporal configuration for sliding windows spans intervals of 4.4 ms, 5 ms, 0.5 s, 1 s, and 1.5 s. In the specific context of the DVS128 Gesture dataset, the DVS Action dataset and the HMBD51-DVS dataset, the temporal overlap between neighboring windows is established at 0.25 s, while in other datasets, this parameter is set to 0.5 s. Given the prevalence of considerable noise within the DVS Action dataset and THU$^{\text{E-ACT}}$-50-CHL dataset during recording, we implemented a denoising method to minimize the likelihood of sampling noise points. The sliding windows of the regressed datasets are all consistent with the frequency of the labels.

\subsubsection{Training Hyperparameter}
Our training model employs the subsequent set of hyperparameters. \textbf{Classification}: Batch Size (32), Number of Points $N$ (1024 or 2048), Optimizer (Adam), Initial Learning Rate (0.001), Scheduler (cosine), and Maximum Epochs (350). \textbf{Regression}: Batch Size (96), Number of Points $N$ (1024), Optimizer (Adam), Initial Learning Rate (0.001), Scheduler (cosine), and Maximum Epochs (1000).
\subsubsection{Evaluation Metrics}
The \textbf{Accuracy} metric used in the classification task represents the ratio of correctly predicted samples to the total number of samples \cite{ren2023spikepoint}. The \textbf{Euclidean distance} is utilized to compare the predicted position with the ground truth in CPR tasks \cite{nguyen2019real}. Positional errors are measured in meters, while orientation errors are measured in degrees. In the eye-tracking tasks, pupil detection accuracy is evaluated based on the Euclidean distance between the predicted pupil center and the ground truth. Detection rates are considered successful if the distance is shorter than \textbf{$p$ pixel} \cite{chen20233et}.
\subsubsection{Network Structure}
All models use a three-stage hierarchy structure, and the number of centroids for each stage is halved so that if the input is 1024 points, the corresponding list of centroids is [512, 256, 128]. On the other hand, as the stage increases, the features of the Point Cloud increase; the feature dimensions are the following three lists: [32, 64, 128], [64, 128, 256], [128, 256, 512]. The member number of the KNN grouping is 24. Both the classifier and regressor are fully connected layers with one hidden layer of 256. The eye-tracking task has an additional sigmoid layer in the final output layer. The number of parameters of the model is directly related to the feature dimension of the features, and the computational FLOPs are directly related to the number of points discovered in Table \ref{table: ablation experiment}.

\begin{table}[t]
\caption{Accuracy and Complexity on DVS128 Gesture.}
\centering
\renewcommand\arraystretch{1}
\scalebox{1}{
\begin{tabular}{cccc}
\hline
Method     & Param.($\text{x}10^6$) & GFLOPs & Acc   \\ \hline
TBR+I3D \cite{innocenti2021temporal}    & 12.25  & 38.82  & 0.996 \\
Event Frames + I3D \cite{bi2020graph} &12.37 &30.11 & 0.965 \\
EV-VGCNN \cite{deng2021ev}   & 0.82   & 0.46 & 0.957 \\
RG-CNN \cite{miao2019neuromorphic}     & 19.46  & 0.79      & 0.961 \\
PointNet++ \cite{wang2019space} & 1.48   & 0.872  & 0.953 \\ 
PLIF \cite{fang2021incorporating} & 1.7   & - & 0.976 \\
GET \cite{peng2023get} & 4.5   & - & 0.979 \\
Swin-T v2 \cite{liu2022swin} & 7.1   & - & 0.932\\\hline
TTPOINT   & 0.334  & 0.587  & 0.988\\ 
\textbf{EventMamba}   & \textbf{0.29}  & \textbf{0.219}  &\textbf{0.992}\\ 
\hline
\end{tabular}
}
\label{table: dvs128}
\end{table}
\subsection{Results of Action Recognition}

\subsubsection{DVS128 Gesture}
Table \ref{table: dvs128} provides a comprehensive overview of the performance and network size metrics for the DVS128 Gesture. The highest recorded accuracy, an impressive 99.6\%, was reported in a frame-based approach \cite{innocenti2021temporal}. However, it's important to note that this method imposes substantial memory and computational demands, with a footprint approximately 42 times larger than our model. In contrast, EventMamba employs temporal feature extraction modules, resulting in a parameter size of a mere 0.29 M and a total number of FLOPs of 0.219 GFLOPs. Remarkably, this efficient model still achieves a commendable accuracy of 99.2\% on the DVS128 Gesture.
\begin{table}[t]
\caption{Accuracy and Complexity on Daily DVS.}
\centering
\renewcommand\arraystretch{1}
\scalebox{1}{
\begin{tabular}{cccc}
\hline
Name              & Param.($\text{x}10^6$) & GFLOPs & Acc    \\ \hline
I3D\cite{carreira2017quo}     & 49.19 &59.28    & 0.962 \\
TANet\cite{liu2021tam}                & 24.8  &65.94    & 0.965 \\
VMV-GCN \cite{xie2022vmv}            & 0.84   &\textbf{0.33 } & 0.941 \\
TimeSformer \cite{bertasius2021space}        & 121.27  &379.7   & 0.906 \\
Motion-based SNN \cite{liu2021event}    & -    & - & 0.903 \\
SpikePoint\cite{ren2023spikepoint}    & 0.16    & - & 0.979 \\\hline
TTPOINT      & \textbf{0.335}  &0.587 & 0.991 \\ 
\textbf{EventMamba}   & 0.905  &0.953 & \textbf{0.991} \\
\hline
\end{tabular}
}
\label{table: daily dvs}
\end{table}
\subsubsection{Daily DVS}
EventMamba outperformed all other methods, achieving the same impressive accuracy of 99.1\% with TTPOINT, as presented in Table \ref{table: daily dvs}.
In stark contrast, the frame-based approach, despite employing nearly 54 times more parameters, yielded a notably lower accuracy. The Event Cloud within this dataset exhibited exceptional recording quality, minimal noise interference, and distinctly discernible actions. Leveraging the Point Cloud sampling method, we are able to precisely capture the intricate motion characteristics of these actions. On the other hand, the model fitting ability becomes stronger, and we increase the point number to avoid the overfitting phenomenon.
\begin{table}[t]
\caption{Accuracy and Complexity on DVS Action.}
\centering
\renewcommand\arraystretch{1}
\scalebox{1}{
\begin{tabular}{cccc}
\hline
Method     & Param.($\text{x}10^6$) &GFLOPs       & Acc   \\ \hline
Deep SNN \cite{gu2019stca}   & -   & -      & 0.712 \\
Motion-based SNN \cite{liu2021event} & -   & -     & 0.781 \\
PointNet  \cite{qi2017pointnet} &3.46 &  \textbf{0.450}     & 0.751 \\
ST-EVNet  \cite{wang2020st}   &1.6 & -    & 0.887 \\ \hline
TTPOINT      & \textbf{0.334}  & 0.587   & 0.927\\
\textbf{EventMamba}      & 0.905  & 0.476   & \textbf{0.891} \\
\hline
\end{tabular}
}
\label{table: dvs action}
\end{table}
\begin{table}[t]
\caption{Accuracy and Complexity  on HMDB51-DVS.}
\centering
\renewcommand\arraystretch{1}
\scalebox{1}{
\begin{tabular}{cccc}
\hline
Method     & Param.($\text{x}10^6$) & GFLOPs & Acc   \\ \hline
C3D \cite{tran2015learning}        & 78.41  & 39.69  & 0.417 \\
I3D \cite{carreira2017quo}      & 12.37  & 30.11  & 0.466 \\
ResNet-34 \cite{he2016deep}  & 63.70  & 11.64  & 0.438 \\
ResNext-50 \cite{hara2018can} & 26.05  & 6.46   & 0.394 \\
ST-ResNet-18 \cite{samadzadeh2020convolutional} & -  & -   & 0.215 \\
RG-CNN \cite{bi2020graph}     & 3.86   & 12.39  & 0.515 \\
EVTC+I3D \cite{xie2022event}     & - & -  & 0.704 \\
EventMix \cite{shen2023eventmix}     & 12.63   & 1.99  & 0.703 \\
\hline
TTPOINT   & \textbf{0.345}  & \textbf{0.587}  & 0.569\\ 
\textbf{EventMamba}   & 0.916 & 0.953 & \textbf{0.864} \\ \hline
\end{tabular}
}
\label{table: hmdb51-dvs}
\end{table}
\begin{table*}
\caption{Camera Pose Relocalization Results. }
\centering
\renewcommand\arraystretch{1}
\scalebox{0.8}{
\begin{tabular}{ccccccc|cc}
\hline
Network             & PoseNet \cite{kendall2015posenet}              & Bayesian PoseNet \cite{kendall2016modelling}     & Pairwise-CNN \cite{laskar2017camera}       & LSTM-Pose \cite{walch2017image}           & SP-LSTM \cite{nguyen2019real}             & CNN-LSTM \cite{tabia2022deep}            & PEPNet \cite{ren2024simple}              & EventMamba            \\ \hline
Parameter           & 12.43M               & 22.35M               & 22.34M               & 16.05M               & 135.25M              & 12.63M               & 0.774M               & 0.904M                \\
FLOPs               & 1.584G               & 3.679G               & 7.359G               & 1.822G               & 15.623G              & 1.998G               & 0.459G               & 0.476G                \\ 
Inference time              & 5.46 ms               & 6.46ms               & 12.46 ms              & 9.95ms               & 5.25ms              & -               & 6.8ms               & 5.5ms               \\
\hline
shapes\_rotation    & 0.109m,7.388$^\circ$ & 0.142m,9.557$^\circ$ & 0.095m,6.332$^\circ$ & 0.032m,4.439$^\circ$ & 0.025m,2.256$^\circ$ & 0.012m,1.652$^\circ$ & 0.005m,1.372$^\circ$ & \textbf{0.004m},\textbf{1.091}$^\circ$ \\
box\_translation    & 0.193m,6.977$^\circ$ & 0.190m,6.636$^\circ$ & 0.178m,6.153$^\circ$ & 0.083m,6.215$^\circ$ & 0.036m,2.195$^\circ$ & \textbf{0.013m},0.873$^\circ$ & 0.017m,0.845$^\circ$ & 0.016m,\textbf{0.810}$^\circ$  \\
shapes\_translation & 0.238m,6.001$^\circ$ & 0.264m,6.235$^\circ$ & 0.201m,5.146$^\circ$ & 0.056m,5.018$^\circ$ & 0.035m,2.117$^\circ$ & 0.020m,1.471$^\circ$ & 0.011m,\textbf{0.582}$^\circ$ & \textbf{0.010m},0.600$^\circ$  \\
dynamic\_6dof       & 0.297m,9.332$^\circ$ & 0.296m,8.963$^\circ$ & 0.245m,5.962$^\circ$ & 0.097m,6.732$^\circ$ & 0.031m,2.047$^\circ$ & 0.016m,1.662$^\circ$ & 0.015m,1.045$^\circ$ & \textbf{0.014m},\textbf{0.911}$^\circ$  \\
hdr\_poster         & 0.282m,8.513$^\circ$ & 0.290m,8.710$^\circ$ & 0.232m,7.234$^\circ$ & 0.108m,6.186$^\circ$ & 0.051m,3.354$^\circ$ & 0.033m,2.421$^\circ$ & 0.016m,0.991$^\circ$ & \textbf{0.015m},\textbf{0.985}$^\circ$  \\
poster\_translation & 0.266m,6.516$^\circ$ & 0.264m,5.459$^\circ$ & 0.211m,6.439$^\circ$ & 0.079m,5.734$^\circ$ & 0.036m,2.074$^\circ$ & 0.020m,1.468$^\circ$ & \textbf{0.012}m,0.588$^\circ$ & 0.013m,\textbf{0.587}$^\circ$ \\ \hline
Average             & 0.231m,7.455$^\circ$ & 0.241m,7.593$^\circ$ & 0.194m,6.211$^\circ$ & 0.076m,5.721$^\circ$ & 0.036m,2.341$^\circ$ & 0.019m,1.591$^\circ$ & 0.013m,0.904$^\circ$ & \textbf{0.012m},\textbf{0.831}$^\circ$  \\ \hline
\end{tabular}}
\label{table:cpr outcome}
\end{table*}
\begin{table*}
\caption{Data Preprocessing Time between Two Methods. }
\centering
\renewcommand\arraystretch{1}
\scalebox{1}{
\begin{tabular}{c|cccccc|c}
\hline
Methods        & shape rotation & box translation & shape translation & dynamic 6dof & hdr poster & poster translation & Average \\ \hline
Sampling event & 0.045 ms        & 0.088 ms         & 0.043 ms           & 0.05 ms       & 0.082 ms    & 0.08 ms             & 0.065 ms \\
Stacking event & 0.38 ms         & 0.53 ms          & 0.37 ms            & 0.45 ms       & 0.48 ms     & 0.53 ms             & 0.457 ms \\ \hline
\end{tabular}}
\label{table:cpr preprocess}
\end{table*}
\begin{table}[t]
\caption{Accuracy and Complexity on UCF101-DVS.}
\centering
\renewcommand\arraystretch{1}
\scalebox{1}{
\begin{tabular}{cccc}
\hline
Name          & Param.($\text{x}10^6$) & GFLOPs & Acc   \\ \hline
C3D \cite{tran2015learning}          & 78.41                         & 39.69  & 0.472 \\
ResNext-50.3D \cite{hara2018can} & 26.05                          & 6.46   & 0.602 \\
I3D \cite{carreira2017quo}          & 12.37                         & 30.11  & 0.635 \\
RG-CNN \cite{bi2020graph}       & 6.95                          & 12.46  & 0.632 \\
ECSNet  \cite{chen2022ecsnet}      & -                             & 12.24  & 0.702 \\ 
EVTC+I3D \cite{xie2022event}     & - & -  & 0.791 \\
Event-LSTM \cite{annamalai2022event} &21.4  &- &0.776 \\ \hline
TTPOINT     & \textbf{0.357}                         & \textbf{0.587}  & 0.725 \\ 
EventMamba   & 3.28 & 3.722 & \textbf{0.979} \\\hline
\end{tabular}
}
\label{table: ucf101}
\end{table}
\begin{table}[t]
\caption{Accuracy and Complexity on THU$^{\text{E-ACT}}$-50-CHL.}
\centering
\renewcommand\arraystretch{1}
\scalebox{1}{
\begin{tabular}{cccc}
\hline
Method     & Param($\text{x}10^6$) & GFLOPs & Acc  \\ \hline
HMAX SNN \cite{xiao2019event}   & -                     & -      & 32.7 \\
Motion-based SNN \cite{liu2021event}& -                     & -      & 47.3 \\
EV-ACT \cite{gao2023action}    & 21.3                  & 29     & 58.5 \\ \hline
\textbf{EventMamba} & \textbf{0.905 }                 & \textbf{0.953 } & \textbf{59.4} \\ \hline
\end{tabular}
}
\label{table: thu}
\end{table}
\subsubsection{DVS Action}
EventMamba achieves an accuracy of 89.1\% on the DVS Action dataset as shown in Table \ref{table: dvs action}. Although the model has a strong ability to capture spatial and temporal features, it is easy to have the risk of overfitting on small datasets, and after tuning down the model and some general methods to solve the overfitting. The model still can not achieve the same high accuracy as TTPOINT, more such as tensor decomposition, distillation, and other techniques may alleviate this problem.

\subsubsection{HMDB51-DVS} The aforementioned datasets are of limited size, and our ongoing efforts aim to ascertain the effectiveness of EventMamba on larger datasets. The HMDB51-DVS dataset comprises 51 distinct action categories.  The data preprocessing settings are delineated in Table \ref{table: infor for dataset}, wherein we adapted the input points number to 2048 and increased the model's dimensionality correspondingly. EventMamba's performance is detailed in the Table \ref{table: hmdb51-dvs}.  While it exhibits a slight increase in parameters and FLOPs compared to the preceding model, it remains lightweight relative to frame-based methodologies. In terms of accuracy, EventMamba achieved state-of-the-art (SOTA) performance, surpassing the previous method \cite{ren2023ttpoint} by 52\%. This finding is particularly noteworthy, given that the HMDB51 dataset, prior to synthesis, currently holds a SOTA accuracy of 88.1\% \cite{wang2023videomae}, with EventMamba demonstrating a close performance of 86.4\%.

\subsubsection{UCF101-DVS}
For the larger 101-class dataset, EventMamba still performs strongly, reaching an impressive 97.1\%, far exceeding all previous methods as shown in Table \ref{table: ucf101}. It is also very close to the 99.6\% of the SOTA method \cite{wang2023videomae} of UCF101 before synthesis.
EVTC and Event-LSTM utilize the huge model and computational resources still only approach 80\% accuracy. This is a great demonstration of the superiority of fine-grained temporal information and explicit temporal features. Additionally, we implement the dataset partitioning method based on filename, achieving accuracies of 60.4\% and 90.3\% for HMDB51-DVS and UCF101-DVS, respectively, still maintaining a leading performance level.

\subsubsection{THU$^{\text{E-ACT}}$-50-CHL}
In more challenging environments, EventMamba demonstrates superior performance over EV-ACT, a method based on multi-representation fusion, while consuming only 15\% of the hardware resources demonstrated in Table \ref{table: thu}. This underscores the compelling evidence supporting the efficacy of Point Cloud representations and our meticulously designed spatio-temporal structure.

In summary, our study employed the EventMamba model to conduct an all-encompassing evaluation of action recognition datasets, ranging from small to large scales. Our analysis demonstrated that EventMamba possesses a remarkable ability to generalize and achieve outstanding performance across datasets of varying sizes.  However, after capturing explicit event temporal features, the model's representational ability is greatly enhanced, and there is a greater risk of overfitting on small datasets, such as DVS Action and Daily DVS.
\begin{table*}
\caption{Event-based Eye Tracking Results. }
\centering
\renewcommand\arraystretch{1}
\scalebox{1}{
\begin{tabular}{ccccccccc}
\hline
Method                           & Resolution & Representation & Param (M) & FLOPs (M) & $p_{3}$ & $p_{5}$ & $p_{10}$ & mse(px) \\ \hline
EventMamba                       & 180 x 240  & P              & 0.903     & 476       & \textbf{0.944}   & \textbf{0.992 }  & \textbf{0.999}    & \underline{1.48}    \\ 
$\text{EventMamba}_{\text{lite}}$                       & 180 x 240  & P              & \underline{0.27}     & 62.6       & \underline{0.936}   &\underline{0.986 } &\underline{ 0.998 }   & \textbf{1.45}  \\ \hline
FAPNet \cite{lin2024fapnet}                          & 180 x 240  & P              & 0.29     & 58.7      & 0.920   & 0.991   & 0.996    & 1.56    \\
PEPNet  \cite{ren2024simple}                         & 180 x 240  & P              & 0.774      & 459       & 0.918   & 0.987   & 0.998    & 1.57   \\ 
$\text{PEPNet}_{\text{tiny}}$ \cite{ren2024simple}     & 180 x 240  & P              & \textbf{0.054}     & 16.25     & 0.786   & 0.945   & 0.995    & 2.2     \\ 
$\text{PointMLP}_{\text{elite}}$ \cite{ma2022rethinking} & 180 x 240  & P              & 0.68      & 924       & 0.840   & 0.977   & 0.997    & 1.96    \\
PointNet++ \cite{qi2017pointnet++}                      & 180 x 240  & P              & 1.46      & 1099      & 0.607   & 0.866   & 0.988    & 3.02    \\
PointNet    \cite{qi2017pointnet}                     & 180 x 240  & P              & 3.46      & 450       & 0.322   & 0.596   & 0.896    & 5.18    \\
CNN    \cite{chen20233et}                          & 60 x 80    & F              & 0.40    & \textbf{18.4}      & 0.578   & 0.774   & 0.914    & -       \\
CB-ConvLSTM  \cite{chen20233et}                     & 60 x 80    & F              & 0.42      & \underline{18.68}     & 0.889   & 0.971   & 0.995    & -       \\
ConvLSTM  \cite{chen20233et}                        & 60 x 80    & F              & 0.42      & 42.61     & 0.887   & 0.971   & 0.994    & -       \\ \hline
\end{tabular}
}
\label{table:eye tracking outcome}
\end{table*}

\begin{table*}
\caption{Ablation Study on HMDB51-DVS Dataset. }
\centering
\renewcommand\arraystretch{1}
\scalebox{0.8}{
\begin{tabular}{c|ccccc}
\hline
\diagbox{Point Number}{Dimension} & [16, 32, 64]            & [32, 64, 128]           & [64, 128, 256]          & 256 w/o Mamba            & 256 w LSTM               \\ \hline
512 points       & (48.2 \%, 0.102 M, 17 M) & (58.6 \%, 0.28 M, 63 M)  & (67.9 \%, 0.92 M, 238 M) & ( 43.9 \%, 0.26 M, 191 M) & ( 58.3 \%, 0.784 M, 229M) \\
1024 points      & (52.7 \%, 0.102 M, 35 M) & (67.8 \%, 0.28 M, 125 M) & (76.6 \%, 0.92 M, 476 M) & ( 52.7 \%, 0.26 M, 383 M) & ( 61.4 \%, 0.784 M, 459M) \\
2048 points      & (59.2 \%, 0.102 M, 71 M) & (74.3 \%, 0.28 M, 252 M) & \textbf{(82.1 \%, 0.92 M, 953 M)} & ( 57.5 \%, 0.26 M, 767 M) & ( 70.8 \%, 0.784 M, 919M) \\ \hline
\end{tabular}
}
\label{table: ablation experiment}
\end{table*}
\begin{table*}[t]
\caption{Abation of Temporal Modules on IJRR Dataset.}
\centering
\renewcommand\arraystretch{1}
\scalebox{1}{
\begin{tabular}{cccc}
\hline
Network             & w/o Mamba   & w LSTM  & EventMamba                    \\ \hline
Parameter           & 0.245M             &0.774M  & 0.904M                        \\
FLOPs               & 0.392G             &0.459G  & 0.476G                        \\ \hline
shapes\_rotation    & 0.007m,1.481$^\circ$ & 0.005m,1.372$^\circ$ & 0.004m,1.091$^\circ$         \\
box\_translation    & 0.019m,1.237$^\circ$ & 0.017m,0.845$^\circ$ & 0.016m,0.810$^\circ$          \\
shapes\_translation & 0.015m,0.827$^\circ$ & 0.011m,0.582$^\circ$ & 0.010m,0.600$^\circ$          \\
dynamic\_6dof       & 0.018m,1.154$^\circ$ & 0.015m,1.045$^\circ$ & 0.014m,0.911$^\circ$          \\
hdr\_poster         & 0.029m,1.744$^\circ$ & 0.016m,0.991$^\circ$ & 0.015m,0.985$^\circ$          \\
poster\_translation & 0.021m,0.991$^\circ$ & 0.012m,0.588$^\circ$ & 0.013m,0.587$^\circ$         \\ \hline
Average             & 0.018m,1.239$^\circ$ & 0.013m,0.904$^\circ$ (+27.3\%) & 0.012m,0.831$^\circ$(+33.1\%) \\ \hline
\end{tabular}
}
\label{table: ablation temproal}
\end{table*}
\subsection{Results of CPR Regression Dataset}
Based on the findings presented in Table \ref{table:cpr outcome}, it is apparent that EventMamba surpasses other models concerning both rotation and translation errors across almost all sequences. Notably, EventMamba achieves these impressive results despite utilizing significantly fewer model parameters and FLOPs compared to the frame-based approach. Compared to the SOTA point-based CPR network PEPNet, the computational resource consumption preserves consistency, inference speed increased by 20\%, and the performance has shown a nearly 8\% improvement. This underscores the pivotal role of explicit temporal feature extraction, exemplified by Mamba, in seamlessly integrating hierarchical structures. Moreover, EventMamba not only exhibits a remarkable 43\% improvement in the average error compared to the SOTA frame-based CNN-LSTM method but also attains superior results across nearly all sequences, surpassing the previous SOTA outcomes. 
In addressing the more intricate and challenging hdr\_poster sequences, while the frame-based approach relies on a denoising network to yield improved results \cite{jin20216}, EventMamba excels by achieving remarkable performance without any additional processing. This observation strongly implies that EventMamba's Point Cloud approach exhibits greater robustness compared to the frame-based method, highlighting its inherent superiority in handling complex scenarios.

We also present in Table \ref{table:cpr preprocess} the time consumed for point-based and frame-based preprocessing. 
Point Clouds exhibit a distinct superiority over other frame-based methods in time consumption for preprocessing (sampling for Point Cloud and stacking events for event frame). We tested the runtime of two preprocessing methods across six sequences on our server. The frame-based methods with the addition of preprocessing time will have a significant performance penalty.
\subsection{Results of Eye Tracking Dataset}
Additionally, we conducted experiments with event-based eye tracking for the regression task. In contrast to the CPR task, which necessitates regression of translation and rotation in the world's coordinate system, the eye tracking task solely requires regression of the pupil's pixel coordinates within the camera coordinate system. To compare EventMamba's performance more extensively, we use several representative Point Cloud networks \cite{ma2022rethinking,qi2017pointnet,qi2017pointnet++,ren2024simple} for benchmarking. Table \ref{table:eye tracking outcome} illustrates that the optimal outcomes are highlighted in bold, whereas the second-best solutions are denoted in the underline. In comparison to classic Point Cloud networks, such as PointNet, PointNet++ and $\text{PointMLP}_{\text{elite}}$, 
EventMamba stands out by significantly enhancing accuracy and reducing distance error, and $\text{EventMamba}_{\text{lite}}$ also maintains high performance while consuming few computational resources. On the other hand, CNN, ConvLSTM, and CB-ConvLSTM eye-tracking frame-based methods \cite{chen20233et} are selected for comparison. Our models exhibit a remarkable improvement in tracking accuracy, with a 3\% increase in the $p_3$ and a 6\% decrease in the MSE error, while maintaining a computational cost comparable to the aforementioned methods. Besides, compared to these frame-based methods, the parameters and FLOPs in EventMamba remain constant regardless of the input camera's resolution. That enables EventMamba to be applied to different resolution edge devices.
\subsection{Inference time}
\begin{table}[t]
\caption{P5, P50, P95 inference time on diverse dataset.}
\centering
\renewcommand\arraystretch{1}
\scalebox{1}{
\begin{tabular}{cccc}
\hline
Dataset    & P5       & P50       & P95       \\ \hline
DVSGesture & 5.6   ms & 7.6   ms  & 9.5   ms  \\
DVSAction  & 5.8   ms & 6.0   ms  & 13.4   ms \\
DailyDVS   & 6.9   ms & 7.2   ms  & 9.2   ms  \\
HMDB51-DVS & 7.1   ms & 8.6   ms  & 15.0   ms \\
UCF101-DVS & 7.3   ms & 14.4   ms & 15.7   ms \\
THU-CHL    & 7.1   ms & 7.3   ms  & 8.4   ms  \\
IJRR       & 5.0   ms & 5.2   ms  & 6.2   ms  \\
3ET        & 5.6   ms & 5.9   ms  & 7.3   ms  \\ \hline
\end{tabular}
}
\label{table: inference time}
\end{table}
We conducted tests on the P5, P50, and P95 inference time of EventMamba across all datasets on our server as shown in Table \ref{table: inference time}. The server is equipped with a GTX 4090 GPU and an AMD 7950X CPU. This Table clearly shows that the inference time for action recognition is significantly shorter than the length of the sliding window, enabling high-speed real-time inference. Additionally, the throughput in eye tracking and camera pose relocation tasks far exceeds the 120 FPS frame rate of RGB cameras, ensuring that the high temporal resolution advantage of event cameras is maintained. 
\subsection{Abation Study}

\subsubsection{Temporal modules}
To demonstrate the effectiveness of temporal modules in the CPR regression task, we replaced it with spatial aggregation and obtained average translation and rotation MSE errors of 0.018m and 1.239 $^\circ$  in the CPR task, which is a 33.1\% decrease compared to the results in the Table \ref{table: ablation temproal}. This table also illustrates that the module responsible for extracting temporal features consumes nearly negligible computational resources yet yields significantly enhanced performance results. On the other hand, Table \ref{table: ablation experiment} demonstrates that the Point Cloud network without a temporal module only achieves 57.5\% accuracy on the HMDB51-DVS dataset. These underscore the critical importance of temporal features in event-based tasks. Temporal information not only conveys the varying actions across different moments in action recognition but also delineates the spatial position of the target within the coordinate system across different temporal axes.
\subsubsection{Point Number}
In contrast to previous work \cite{ren2023ttpoint,ren2023spikepoint}, we increased the number of input points to 2048 on large action recognition datasets, and to determine this significant impact factor, we verified it on HMDB51-DVS. The experiments compare the accuracy at training 100 epoch as described in Table \ref{table: ablation experiment}, EventMamba with 2048 points has 82.1\% accuracy on the testset, with 1024 points has 76.6\% accuracy on the testset, and with 512 points has 67.9\% accuracy on the testset. This shows that increasing the number of input points can significantly improve the performance of the model on large action recognition datasets.
\subsubsection{Mamba or LSTM?}
Table \ref{table: ablation experiment} and \ref{table: ablation temproal} show the performance of the two explicit feature extractors on the HMDB51-DVS and IJRR datasets, respectively. We try to ensure that both models consume an equitable computational resource, although the Mamba-based model still slightly surpasses the LSTM-based model. On the HMDB51-DVS dataset, Mamba demonstrated a notable performance advantage, achieving an accuracy of 82.1\%, which exceeded LSTM's accuracy of 70.8\% by 16\%. Likewise, on the IJRR dataset, Mamba exhibited superior performance with its average Mean Squared Error (MSE) outperforming LSTM's by 8\%. Additionally, the Mamba-based model's inference is faster than LSTM, at 5.5 ms compared to 6.8 ms.
Ignoring the slight increase in computational resources, we believe that Mamba is superior to LSTM in processing sequence events data.
\section{Conclusion}
In this paper, we rethink the temporal distinction between Event Cloud and Point Cloud and introduce a spatio-temporal hierarchical structure to address various event-based tasks. This innovative paradigm aligns with the inherently sparse and asynchronous characteristics of the event camera and effectively extracts implicit and explicit temporal information between events. Comprehensive experiments substantiate the efficacy of this approach in both regression and classification tasks, and we will derive this to more event-based tasks in the future.
\section{Acknowledgment}
This work was supported in part by the Young Scientists Fund of the National Natural Science Foundation of China (Grant 62305278), as well as the Guangdong Basic and Applied Basic Research Foundation (NO.2025A1515011758).

\ifCLASSOPTIONcaptionsoff
  \newpage
\fi



\bibliographystyle{IEEEtran}
\bibliography{eventmamba}

\begin{thebibliography}{10}
\providecommand{\url}[1]{#1}
\csname url@samestyle\endcsname
\providecommand{\newblock}{\relax}
\providecommand{\bibinfo}[2]{#2}
\providecommand{\BIBentrySTDinterwordspacing}{\spaceskip=0pt\relax}
\providecommand{\BIBentryALTinterwordstretchfactor}{4}
\providecommand{\BIBentryALTinterwordspacing}{\spaceskip=\fontdimen2\font plus
\BIBentryALTinterwordstretchfactor\fontdimen3\font minus \fontdimen4\font\relax}
\providecommand{\BIBforeignlanguage}[2]{{%
\expandafter\ifx\csname l@#1\endcsname\relax
\typeout{** WARNING: IEEEtran.bst: No hyphenation pattern has been}%
\typeout{** loaded for the language `#1'. Using the pattern for}%
\typeout{** the default language instead.}%
\else
\language=\csname l@#1\endcsname
\fi
#2}}
\providecommand{\BIBdecl}{\relax}
\BIBdecl

\bibitem{lichtsteiner2008128}
P.~Lichtsteiner, C.~Posch, and T.~Delbruck, ``A 128 $\times$ 128 120 db 15 $\mu$s latency asynchronous temporal contrast vision sensor,'' \emph{IEEE journal of solid-state circuits}, vol.~43, no.~2, pp. 566--576, 2008.

\bibitem{yao2021temporal}
M.~Yao, H.~Gao, G.~Zhao, D.~Wang, Y.~Lin, Z.~Yang, and G.~Li, ``Temporal-wise attention spiking neural networks for event streams classification,'' in \emph{Proceedings of the IEEE/CVF International Conference on Computer Vision}, 2021, pp. 10\,221--10\,230.

\bibitem{posch2014retinomorphic}
C.~Posch, T.~Serrano-Gotarredona, B.~Linares-Barranco, and T.~Delbruck, ``Retinomorphic event-based vision sensors: bioinspired cameras with spiking output,'' \emph{Proceedings of the IEEE}, vol. 102, no.~10, pp. 1470--1484, 2014.

\bibitem{gallego2020event}
G.~Gallego, T.~Delbr{\"u}ck, G.~Orchard, C.~Bartolozzi, B.~Taba, A.~Censi, S.~Leutenegger, A.~J. Davison, J.~Conradt, K.~Daniilidis \emph{et~al.}, ``Event-based vision: A survey,'' \emph{IEEE transactions on pattern analysis and machine intelligence}, vol.~44, no.~1, pp. 154--180, 2020.

\bibitem{gehrig2019end}
D.~Gehrig, A.~Loquercio, K.~G. Derpanis, and D.~Scaramuzza, ``End-to-end learning of representations for asynchronous event-based data,'' in \emph{Proceedings of the IEEE/CVF International Conference on Computer Vision}, 2019, pp. 5633--5643.

\bibitem{lin2020efficient}
S.~Lin, F.~Xu, X.~Wang, W.~Yang, and L.~Yu, ``Efficient spatial-temporal normalization of sae representation for event camera,'' \emph{IEEE Robotics and Automation Letters}, vol.~5, no.~3, pp. 4265--4272, 2020.

\bibitem{wang2019space}
Q.~Wang, Y.~Zhang, J.~Yuan, and Y.~Lu, ``Space-time event clouds for gesture recognition: From rgb cameras to event cameras,'' in \emph{2019 IEEE Winter Conference on Applications of Computer Vision (WACV)}.\hskip 1em plus 0.5em minus 0.4em\relax IEEE, 2019, pp. 1826--1835.

\bibitem{ren2023ttpoint}
H.~Ren, Y.~Zhou, H.~Fu, Y.~Huang, R.~Xu, and B.~Cheng, ``Ttpoint: A tensorized point cloud network for lightweight action recognition with event cameras,'' in \emph{Proceedings of the 31st ACM International Conference on Multimedia}, 2023, pp. 8026--8034.

\bibitem{ren2024simple}
H.~Ren, J.~Zhu, Y.~Zhou, H.~Fu, Y.~Huang, and B.~Cheng, ``A simple and effective point-based network for event camera 6-dofs pose relocalization,'' \emph{arXiv preprint arXiv:2403.19412}, 2024.

\bibitem{guo2020deep}
Y.~Guo, H.~Wang, Q.~Hu, H.~Liu, L.~Liu, and M.~Bennamoun, ``Deep learning for 3d point clouds: A survey,'' \emph{IEEE transactions on pattern analysis and machine intelligence}, vol.~43, no.~12, pp. 4338--4364, 2020.

\bibitem{qi2017pointnet}
C.~R. Qi, H.~Su, K.~Mo, and L.~J. Guibas, ``Pointnet: Deep learning on point sets for 3d classification and segmentation,'' in \emph{Proceedings of the IEEE conference on computer vision and pattern recognition}, 2017, pp. 652--660.

\bibitem{qi2017pointnet++}
C.~R. Qi, L.~Yi, H.~Su, and L.~J. Guibas, ``Pointnet++: Deep hierarchical feature learning on point sets in a metric space,'' \emph{Advances in neural information processing systems}, vol.~30, 2017.

\bibitem{innocenti2021temporal}
S.~U. Innocenti, F.~Becattini, F.~Pernici, and A.~Del~Bimbo, ``Temporal binary representation for event-based action recognition,'' in \emph{2020 25th International Conference on Pattern Recognition (ICPR)}.\hskip 1em plus 0.5em minus 0.4em\relax IEEE, 2021, pp. 10\,426--10\,432.

\bibitem{annamalai2022event}
L.~Annamalai, V.~Ramanathan, and C.~S. Thakur, ``Event-lstm: An unsupervised and asynchronous learning-based representation for event-based data,'' \emph{IEEE Robotics and Automation Letters}, vol.~7, no.~2, pp. 4678--4685, 2022.

\bibitem{amir2017low}
A.~Amir, B.~Taba, D.~Berg, T.~Melano, J.~McKinstry, C.~Di~Nolfo, T.~Nayak, A.~Andreopoulos, G.~Garreau, M.~Mendoza \emph{et~al.}, ``A low power, fully event-based gesture recognition system,'' in \emph{Proceedings of the IEEE conference on computer vision and pattern recognition}, 2017, pp. 7243--7252.

\bibitem{bi2020graph}
Y.~Bi, A.~Chadha, A.~Abbas, E.~Bourtsoulatze, and Y.~Andreopoulos, ``Graph-based spatio-temporal feature learning for neuromorphic vision sensing,'' \emph{IEEE Transactions on Image Processing}, vol.~29, pp. 9084--9098, 2020.

\bibitem{deng2021learning}
Y.~Deng, H.~Chen, H.~Chen, and Y.~Li, ``Learning from images: A distillation learning framework for event cameras,'' \emph{IEEE Transactions on Image Processing}, vol.~30, pp. 4919--4931, 2021.

\bibitem{miao2019neuromorphic}
S.~Miao, G.~Chen, X.~Ning, Y.~Zi, K.~Ren, Z.~Bing, and A.~Knoll, ``Neuromorphic vision datasets for pedestrian detection, action recognition, and fall detection,'' \emph{Frontiers in neurorobotics}, vol.~13, p.~38, 2019.

\bibitem{kendall2015posenet}
A.~Kendall, M.~Grimes, and R.~Cipolla, ``Posenet: A convolutional network for real-time 6-dof camera relocalization,'' in \emph{Proceedings of the IEEE international conference on computer vision}, 2015, pp. 2938--2946.

\bibitem{shavit2019introduction}
Y.~Shavit and R.~Ferens, ``Introduction to camera pose estimation with deep learning,'' \emph{arXiv preprint arXiv:1907.05272}, 2019.

\bibitem{walch2017image}
F.~Walch, C.~Hazirbas, L.~Leal-Taixe, T.~Sattler, S.~Hilsenbeck, and D.~Cremers, ``Image-based localization using lstms for structured feature correlation,'' in \emph{Proceedings of the IEEE International Conference on Computer Vision}, 2017, pp. 627--637.

\bibitem{naseer2017deep}
T.~Naseer and W.~Burgard, ``Deep regression for monocular camera-based 6-dof global localization in outdoor environments,'' in \emph{2017 IEEE/RSJ International Conference on Intelligent Robots and Systems (IROS)}.\hskip 1em plus 0.5em minus 0.4em\relax IEEE, 2017, pp. 1525--1530.

\bibitem{wu2017delving}
J.~Wu, L.~Ma, and X.~Hu, ``Delving deeper into convolutional neural networks for camera relocalization,'' in \emph{2017 IEEE International Conference on Robotics and Automation (ICRA)}.\hskip 1em plus 0.5em minus 0.4em\relax IEEE, 2017, pp. 5644--5651.

\bibitem{brachmann2021visual}
E.~Brachmann and C.~Rother, ``Visual camera re-localization from rgb and rgb-d images using dsac,'' \emph{IEEE transactions on pattern analysis and machine intelligence}, vol.~44, no.~9, pp. 5847--5865, 2021.

\bibitem{nguyen2019real}
A.~Nguyen, T.-T. Do, D.~G. Caldwell, and N.~G. Tsagarakis, ``Real-time 6dof pose relocalization for event cameras with stacked spatial lstm networks,'' in \emph{Proceedings of the IEEE/CVF Conference on Computer Vision and Pattern Recognition Workshops}, 2019, pp. 0--0.

\bibitem{jin20216}
Y.~Jin, L.~Yu, G.~Li, and S.~Fei, ``A 6-dofs event-based camera relocalization system by cnn-lstm and image denoising,'' \emph{Expert Systems with Applications}, vol. 170, p. 114535, 2021.

\bibitem{lin20226}
H.~Lin, M.~Li, Q.~Xia, Y.~Fei, B.~Yin, and X.~Yang, ``6-dof pose relocalization for event cameras with entropy frame and attention networks,'' in \emph{The 18th ACM SIGGRAPH International Conference on Virtual-Reality Continuum and its Applications in Industry}, 2022, pp. 1--8.

\bibitem{mitrokhin2020learning}
A.~Mitrokhin, Z.~Hua, C.~Fermuller, and Y.~Aloimonos, ``Learning visual motion segmentation using event surfaces,'' in \emph{Proceedings of the IEEE/CVF Conference on Computer Vision and Pattern Recognition}, 2020, pp. 14\,414--14\,423.

\bibitem{ryan2021real}
C.~Ryan, B.~O’Sullivan, A.~Elrasad, A.~Cahill, J.~Lemley, P.~Kielty, C.~Posch, and E.~Perot, ``Real-time face \& eye tracking and blink detection using event cameras,'' \emph{Neural Networks}, vol. 141, pp. 87--97, 2021.

\bibitem{stoffregen2022event}
T.~Stoffregen, H.~Daraei, C.~Robinson, and A.~Fix, ``Event-based kilohertz eye tracking using coded differential lighting,'' in \emph{Proceedings of the IEEE/CVF Winter Conference on Applications of Computer Vision}, 2022, pp. 2515--2523.

\bibitem{angelopoulos2021event}
A.~N. Angelopoulos, J.~N. Martel, A.~P. Kohli, J.~Conradt, and G.~Wetzstein, ``Event-based near-eye gaze tracking beyond 10,000 hz,'' \emph{IEEE Transactions on Visualization and Computer Graphics}, vol.~27, no.~5, pp. 2577--2586, 2021.

\bibitem{zhao2024ev}
G.~Zhao, Y.~Yang, J.~Liu, N.~Chen, Y.~Shen, H.~Wen, and G.~Lan, ``Ev-eye: Rethinking high-frequency eye tracking through the lenses of event cameras,'' \emph{Advances in Neural Information Processing Systems}, vol.~36, 2024.

\bibitem{chen20233et}
Q.~Chen, Z.~Wang, S.-C. Liu, and C.~Gao, ``3et: Efficient event-based eye tracking using a change-based convlstm network,'' in \emph{2023 IEEE Biomedical Circuits and Systems Conference (BioCAS)}.\hskip 1em plus 0.5em minus 0.4em\relax IEEE, 2023, pp. 1--5.

\bibitem{zubic2024state}
N.~Zubi{\'c}, M.~Gehrig, and D.~Scaramuzza, ``State space models for event cameras,'' \emph{arXiv preprint arXiv:2402.15584}, 2024.

\bibitem{wu2019pointconv}
W.~Wu, Z.~Qi, and L.~Fuxin, ``Pointconv: Deep convolutional networks on 3d point clouds,'' in \emph{Proceedings of the IEEE/CVF Conference on Computer Vision and Pattern Recognition}, 2019, pp. 9621--9630.

\bibitem{zhao2021point}
H.~Zhao, L.~Jiang, J.~Jia, P.~H. Torr, and V.~Koltun, ``Point transformer,'' in \emph{Proceedings of the IEEE/CVF International Conference on Computer Vision}, 2021, pp. 16\,259--16\,268.

\bibitem{liang2024pointmamba}
D.~Liang, X.~Zhou, X.~Wang, X.~Zhu, W.~Xu, Z.~Zou, X.~Ye, and X.~Bai, ``Pointmamba: A simple state space model for point cloud analysis,'' \emph{arXiv preprint arXiv:2402.10739}, 2024.

\bibitem{zhang2024point}
T.~Zhang, X.~Li, H.~Yuan, S.~Ji, and S.~Yan, ``Point could mamba: Point cloud learning via state space model,'' \emph{arXiv preprint arXiv:2403.00762}, 2024.

\bibitem{liu2024point}
J.~Liu, R.~Yu, Y.~Wang, Y.~Zheng, T.~Deng, W.~Ye, and H.~Wang, ``Point mamba: A novel point cloud backbone based on state space model with octree-based ordering strategy,'' \emph{arXiv preprint arXiv:2403.06467}, 2024.

\bibitem{ma2022rethinking}
X.~Ma, C.~Qin, H.~You, H.~Ran, and Y.~Fu, ``Rethinking network design and local geometry in point cloud: A simple residual mlp framework,'' \emph{arXiv preprint arXiv:2202.07123}, 2022.

\bibitem{sekikawa2019eventnet}
Y.~Sekikawa, K.~Hara, and H.~Saito, ``Eventnet: Asynchronous recursive event processing,'' in \emph{Proceedings of the IEEE/CVF Conference on Computer Vision and Pattern Recognition}, 2019, pp. 3887--3896.

\bibitem{yang2019modeling}
J.~Yang, Q.~Zhang, B.~Ni, L.~Li, J.~Liu, M.~Zhou, and Q.~Tian, ``Modeling point clouds with self-attention and gumbel subset sampling,'' in \emph{Proceedings of the IEEE/CVF conference on computer vision and pattern recognition}, 2019, pp. 3323--3332.

\bibitem{xie2022vmv}
B.~Xie, Y.~Deng, Z.~Shao, H.~Liu, and Y.~Li, ``Vmv-gcn: Volumetric multi-view based graph cnn for event stream classification,'' \emph{IEEE Robotics and Automation Letters}, vol.~7, no.~2, pp. 1976--1983, 2022.

\bibitem{gu2021efficiently}
A.~Gu, K.~Goel, and C.~R{\'e}, ``Efficiently modeling long sequences with structured state spaces,'' \emph{arXiv preprint arXiv:2111.00396}, 2021.

\bibitem{gupta2022diagonal}
A.~Gupta, A.~Gu, and J.~Berant, ``Diagonal state spaces are as effective as structured state spaces,'' \emph{Advances in Neural Information Processing Systems}, vol.~35, pp. 22\,982--22\,994, 2022.

\bibitem{gu2022parameterization}
A.~Gu, K.~Goel, A.~Gupta, and C.~R{\'e}, ``On the parameterization and initialization of diagonal state space models,'' \emph{Advances in Neural Information Processing Systems}, vol.~35, pp. 35\,971--35\,983, 2022.

\bibitem{gu2023mamba}
A.~Gu and T.~Dao, ``Mamba: Linear-time sequence modeling with selective state spaces,'' \emph{arXiv preprint arXiv:2312.00752}, 2023.

\bibitem{nguyen2022s4nd}
E.~Nguyen, K.~Goel, A.~Gu, G.~Downs, P.~Shah, T.~Dao, S.~Baccus, and C.~R{\'e}, ``S4nd: Modeling images and videos as multidimensional signals with state spaces,'' \emph{Advances in neural information processing systems}, vol.~35, pp. 2846--2861, 2022.

\bibitem{zhu2024vision}
L.~Zhu, B.~Liao, Q.~Zhang, X.~Wang, W.~Liu, and X.~Wang, ``Vision mamba: Efficient visual representation learning with bidirectional state space model,'' \emph{arXiv preprint arXiv:2401.09417}, 2024.

\bibitem{wang2024state}
X.~Wang, S.~Wang, Y.~Ding, Y.~Li, W.~Wu, Y.~Rong, W.~Kong, J.~Huang, S.~Li, H.~Yang \emph{et~al.}, ``State space model for new-generation network alternative to transformers: A survey,'' \emph{arXiv preprint arXiv:2404.09516}, 2024.

\bibitem{huang2024mamba}
J.~Huang, S.~Wang, S.~Wang, Z.~Wu, X.~Wang, and B.~Jiang, ``Mamba-fetrack: Frame-event tracking via state space model,'' \emph{arXiv preprint arXiv:2404.18174}, 2024.

\bibitem{yuan2024learning}
D.~Yuan, L.~Burner, J.~Wu, M.~Liu, J.~Chen, Y.~Aloimonos, and C.~Ferm{\"u}ller, ``Learning normal flow directly from event neighborhoods,'' \emph{arXiv preprint arXiv:2412.11284}, 2024.

\bibitem{liu2021event}
Q.~Liu, D.~Xing, H.~Tang, D.~Ma, and G.~Pan, ``Event-based action recognition using motion information and spiking neural networks.'' in \emph{IJCAI}, 2021, pp. 1743--1749.

\bibitem{gao2023action}
Y.~Gao, J.~Lu, S.~Li, N.~Ma, S.~Du, Y.~Li, and Q.~Dai, ``Action recognition and benchmark using event cameras,'' \emph{IEEE Transactions on Pattern Analysis and Machine Intelligence}, 2023.

\bibitem{mueggler2017event}
E.~Mueggler, H.~Rebecq, G.~Gallego, T.~Delbruck, and D.~Scaramuzza, ``The event-camera dataset and simulator: Event-based data for pose estimation, visual odometry, and slam,'' \emph{The International Journal of Robotics Research}, vol.~36, no.~2, pp. 142--149, 2017.

\bibitem{tonsen2016labelled}
M.~Tonsen, X.~Zhang, Y.~Sugano, and A.~Bulling, ``Labelled pupils in the wild: a dataset for studying pupil detection in unconstrained environments,'' in \emph{Proceedings of the ninth biennial ACM symposium on eye tracking research \& applications}, 2016, pp. 139--142.

\bibitem{hu2021v2e}
Y.~Hu, S.-C. Liu, and T.~Delbruck, ``v2e: From video frames to realistic dvs events,'' in \emph{Proceedings of the IEEE/CVF Conference on Computer Vision and Pattern Recognition}, 2021, pp. 1312--1321.

\bibitem{ren2023spikepoint}
H.~Ren, Y.~Zhou, F.~Haotian, Y.~Huang, L.~Xiaopeng, J.~Song, and B.~Cheng, ``Spikepoint: An efficient point-based spiking neural network for event cameras action recognition,'' in \emph{The Twelfth International Conference on Learning Representations}, 2023.

\bibitem{deng2021ev}
Y.~Deng, H.~Chen, H.~Chen, and Y.~Li, ``Ev-vgcnn: A voxel graph cnn for event-based object classification,'' \emph{arXiv preprint arXiv:2106.00216}, 2021.

\bibitem{fang2021incorporating}
W.~Fang, Z.~Yu, Y.~Chen, T.~Masquelier, T.~Huang, and Y.~Tian, ``Incorporating learnable membrane time constant to enhance learning of spiking neural networks,'' in \emph{Proceedings of the IEEE/CVF international conference on computer vision}, 2021, pp. 2661--2671.

\bibitem{peng2023get}
Y.~Peng, Y.~Zhang, Z.~Xiong, X.~Sun, and F.~Wu, ``Get: group event transformer for event-based vision,'' in \emph{Proceedings of the IEEE/CVF International Conference on Computer Vision}, 2023, pp. 6038--6048.

\bibitem{liu2022swin}
Z.~Liu, H.~Hu, Y.~Lin, Z.~Yao, Z.~Xie, Y.~Wei, J.~Ning, Y.~Cao, Z.~Zhang, L.~Dong \emph{et~al.}, ``Swin transformer v2: Scaling up capacity and resolution,'' in \emph{Proceedings of the IEEE/CVF conference on computer vision and pattern recognition}, 2022, pp. 12\,009--12\,019.

\bibitem{carreira2017quo}
J.~Carreira and A.~Zisserman, ``Quo vadis, action recognition? a new model and the kinetics dataset,'' in \emph{proceedings of the IEEE Conference on Computer Vision and Pattern Recognition}, 2017, pp. 6299--6308.

\bibitem{liu2021tam}
Z.~Liu, L.~Wang, W.~Wu, C.~Qian, and T.~Lu, ``Tam: Temporal adaptive module for video recognition,'' in \emph{Proceedings of the IEEE/CVF international conference on computer vision}, 2021, pp. 13\,708--13\,718.

\bibitem{bertasius2021space}
G.~Bertasius, H.~Wang, and L.~Torresani, ``Is space-time attention all you need for video understanding?'' in \emph{ICML}, vol.~2, no.~3, 2021, p.~4.

\bibitem{gu2019stca}
P.~Gu, R.~Xiao, G.~Pan, and H.~Tang, ``Stca: Spatio-temporal credit assignment with delayed feedback in deep spiking neural networks.'' in \emph{IJCAI}, 2019, pp. 1366--1372.

\bibitem{wang2020st}
Q.~Wang, Y.~Zhang, J.~Yuan, and Y.~Lu, ``‘st-evnet: Hierarchical spatial and temporal feature learning on space-time event clouds,'' \emph{Proc. Adv. Neural Inf. Process. Syst.(NeurlIPS)}, 2020.

\bibitem{tran2015learning}
D.~Tran, L.~Bourdev, R.~Fergus, L.~Torresani, and M.~Paluri, ``Learning spatiotemporal features with 3d convolutional networks,'' in \emph{Proceedings of the IEEE international conference on computer vision}, 2015, pp. 4489--4497.

\bibitem{he2016deep}
K.~He, X.~Zhang, S.~Ren, and J.~Sun, ``Deep residual learning for image recognition,'' in \emph{Proceedings of the IEEE conference on computer vision and pattern recognition}, 2016, pp. 770--778.

\bibitem{hara2018can}
K.~Hara, H.~Kataoka, and Y.~Satoh, ``Can spatiotemporal 3d cnns retrace the history of 2d cnns and imagenet?'' in \emph{Proceedings of the IEEE conference on Computer Vision and Pattern Recognition}, 2018, pp. 6546--6555.

\bibitem{samadzadeh2020convolutional}
A.~Samadzadeh, F.~S.~T. Far, A.~Javadi, A.~Nickabadi, and M.~H. Chehreghani, ``Convolutional spiking neural networks for spatio-temporal feature extraction,'' \emph{arXiv preprint arXiv:2003.12346}, 2020.

\bibitem{xie2022event}
B.~Xie, Y.~Deng, Z.~Shao, H.~Liu, Q.~Xu, and Y.~Li, ``Event tubelet compressor: Generating compact representations for event-based action recognition,'' in \emph{2022 7th International Conference on Control, Robotics and Cybernetics (CRC)}.\hskip 1em plus 0.5em minus 0.4em\relax IEEE, 2022, pp. 12--16.

\bibitem{shen2023eventmix}
G.~Shen, D.~Zhao, and Y.~Zeng, ``Eventmix: An efficient data augmentation strategy for event-based learning,'' \emph{Information Sciences}, vol. 644, p. 119170, 2023.

\bibitem{kendall2016modelling}
A.~Kendall and R.~Cipolla, ``Modelling uncertainty in deep learning for camera relocalization,'' in \emph{2016 IEEE international conference on Robotics and Automation (ICRA)}.\hskip 1em plus 0.5em minus 0.4em\relax IEEE, 2016, pp. 4762--4769.

\bibitem{laskar2017camera}
Z.~Laskar, I.~Melekhov, S.~Kalia, and J.~Kannala, ``Camera relocalization by computing pairwise relative poses using convolutional neural network,'' in \emph{Proceedings of the IEEE International Conference on Computer Vision Workshops}, 2017, pp. 929--938.

\bibitem{tabia2022deep}
A.~Tabia, F.~Bonardi, and S.~Bouchafa, ``Deep learning for pose estimation from event camera,'' in \emph{2022 International Conference on Digital Image Computing: Techniques and Applications (DICTA)}.\hskip 1em plus 0.5em minus 0.4em\relax IEEE, 2022, pp. 1--7.

\bibitem{chen2022ecsnet}
Z.~Chen, J.~Wu, J.~Hou, L.~Li, W.~Dong, and G.~Shi, ``Ecsnet: Spatio-temporal feature learning for event camera,'' \emph{IEEE Transactions on Circuits and Systems for Video Technology}, 2022.

\bibitem{xiao2019event}
R.~Xiao, H.~Tang, Y.~Ma, R.~Yan, and G.~Orchard, ``An event-driven categorization model for aer image sensors using multispike encoding and learning,'' \emph{IEEE transactions on neural networks and learning systems}, vol.~31, no.~9, pp. 3649--3657, 2019.

\bibitem{wang2023videomae}
L.~Wang, B.~Huang, Z.~Zhao, Z.~Tong, Y.~He, Y.~Wang, Y.~Wang, and Y.~Qiao, ``Videomae v2: Scaling video masked autoencoders with dual masking,'' in \emph{Proceedings of the IEEE/CVF Conference on Computer Vision and Pattern Recognition}, 2023, pp. 14\,549--14\,560.

\bibitem{lin2024fapnet}
X.~Lin, H.~Ren, and B.~Cheng, ``{FAPN}et: {A}n {E}ffective {F}requency {A}daptive {P}oint-based {E}ye {T}racker,'' in \emph{Proceedings of the IEEE/CVF Conference on Computer Vision and Pattern Recognition Workshops}, 2024.

\end{thebibliography}

\end{document}